\RequirePackage{fix-cm}
\documentclass[twocolumn]{svjour3} 
\usepackage{natbib}
\smartqed
\usepackage{graphicx}
\usepackage{amsmath}
\usepackage{amssymb} % define this before the line numbering.
\usepackage{color}
\usepackage{xcolor}
\usepackage{hyperref}

\usepackage{multirow}
\usepackage{subfig}
\usepackage{gensymb}
\newcommand{\tabincell}[2]{\begin{tabular}{@{}#1@{}}#2\end{tabular}}

\newcommand{\dimnet}[0]{DeepIM}
\newcommand{\xreal}[0]{\mathbf{img}_{\text{obs}}}
\newcommand{\xrend}[0]{\mathbf{img}_{\text{rend}}}
\newcommand{\mreal}[0]{\mathbf{m}_{\text{obs}}}
\newcommand{\mrend}[0]{\mathbf{m}_{\text{rend}}}
\newcommand{\rend}[0]{rendered}
\newcommand{\real}[0]{observed}
\newcommand{\pose}[0]{\mathbf{p}}

\newcommand{\gtpose}[0]{\mathbf{\hat{p}}}

\newcommand{\yi}[1]{#1}

\graphicspath{{figs/}}
\begin{document}
%\linenumbers
% \pagestyle{headings}
% \mainmatter
% \def\ECCV18SubNumber{1903}  % Insert your submission number here

\title{DeepIM: Deep Iterative Matching for 6D Pose Estimation} % Replace with your title

%\orcidID{0000-0002-7547-5073}

\author{Yi Li \and Gu Wang \and Xiangyang Ji \and Yu Xiang \and Dieter Fox}
% \authorrunning{Yi Li, Gu Wang, Xiangyang Ji, Yu Xiang, Dieter Fox}
\institute{Yi Li \at 
           University of Washington, Tsinghua University and BNRist \\
           \email{yili.matrix@gmail.com} \\
           Gu Wang \at
           Tsinghua University and BNRist \\
           \email{wangg16@mails.tsinghua.edu.cn} \\
           Xiangyang Ji \at 
           Tsinghua University and BNRist \\
           \email{xyji@tsinghua.edu.cn} \\
           Yu Xiang \at 
           NVIDIA \\
           \email{yux@nvidia.com} \\
           Dieter Fox \at 
           University of Washington and NVIDIA \\
           \email{dieterf@nvidia.com}} 

\maketitle

\begin{abstract}

Estimating 6D poses of objects from images is an important problem in various applications such as robot manipulation and virtual reality. While direct regression of images to object poses has limited accuracy, matching rendered images of an object against the input image can produce accurate results. In this work, we propose a novel deep neural network for 6D pose matching named \dimnet. Given an initial pose estimation, our network is able to iteratively refine the pose by matching the \rend\ image against the observed image. The network is trained to predict a relative pose transformation using a disentangled representation of 3D location and 3D orientation and an iterative training process. Experiments on two commonly used benchmarks for 6D pose estimation demonstrate that \dimnet \ achieves large improvements over state-of-the-art methods. We furthermore show that \dimnet\ is able to match previously unseen objects.

\end{abstract}

\section{Introduction}

Localizing objects in 3D from images is important in many real world applications. For instance, in a robot manipulation task, the ability to recognize the 6D pose of objects, i.e., 3D location and 3D orientation of objects, provides useful information for grasp and motion planning. In a virtual reality application, 6D object pose estimation enables virtual interactions between human and objects. While several recent techniques have used depth cameras
for object pose estimation, such cameras have limitations with respect to frame rate, field of view, resolution, and depth range, making it very difficult to detect small, thin, transparent, or fast moving objects. Unfortunately, RGB-only 6D object pose estimation is still a challenging problem, since the appearance of objects in the images changes according to a number of factors, such as lighting, pose variations, and occlusions between objects. Furthermore, a robust 6D pose estimation method needs to handle both textured and textureless objects. 

\begin{figure*}[t]
	\centering
	\includegraphics[width = \linewidth]{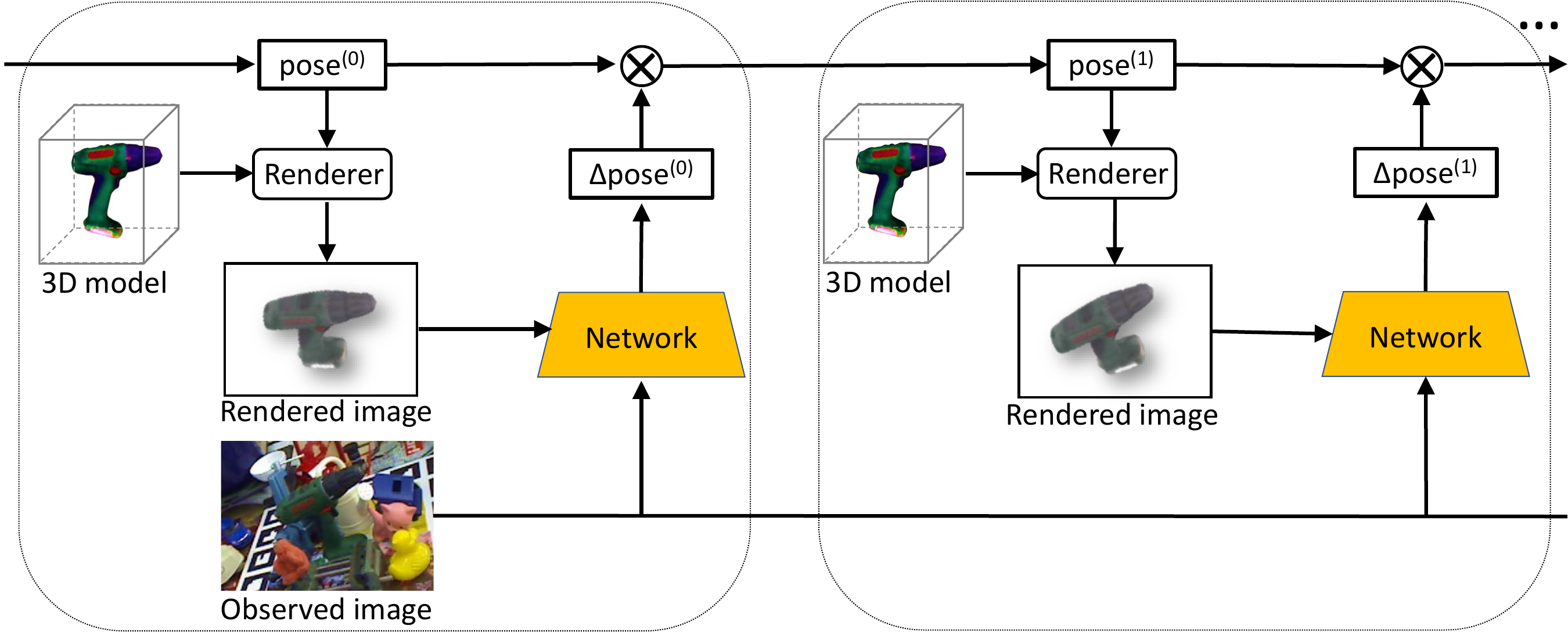}
	\caption{We propose \dimnet, a deep iterative matching network for 6D object pose estimation. The network is trained to predict a relative SE(3) transformation that can be applied to an initial pose estimation for iterative pose refinement. Given a 6D pose estimation of an object, which can be the output of other pose estimation methods like PoseCNN~\citep{xiang2017posecnn} ($\text{pose}^{(0)}$ in the figure) or the refined pose from previous iteration ($\text{pose}^{(1)}$ in the figure), along with the 3D model of the object, we generate the rendered image showing the appearance of the target object under this rough pose estimation. With the image pairs of rendered image and observed image, the network predicts a relative transformation ($\mathrm{\Delta}\text{pose}$ in the figure) which can be applied to refine the input pose. The refined pose can be used as the input pose of next iteration and therefore the process can be repeated until the refined pose converges or the number of iterations reaches a pre-determined number.}
	\label{fig:intro}
\end{figure*}

Traditionally, the 6D pose estimation problem has been tackled by matching local features extracted from an image to features in a 3D model of the object~\citep{lowe1999object,rothganger20063d,collet2011moped}. By using the 2D-3D correspondences, the 6D pose of the object can be recovered. Unfortunately, such methods cannot handle textureless objects well since only few local features can be extracted for them. To handle textureless objects, two classes of approaches were proposed in the literature. Methods in the first class learn to estimate the 3D model coordinates of pixels or keypoints of the object in the input image. In this way, the 2D-3D correspondences are established for 6D pose estimation~\citep{Brachmann2014Learning6O,rad2017bb8,tekin2017real}. Methods in the second class convert the 6D pose estimation problem into a pose classification problem by discretizing the pose space~\citep{hinterstoisser2012accv} or into a pose regression problem~\citep{xiang2017posecnn}. These methods can deal with textureless objects, but they are not able to achieve highly accurate pose estimation, since small errors in the classification or regression stage directly lead to pose mismatches. A common way to improve the pose accuracy is pose refinement: Given an initial pose estimation, a synthetic RGB image can be rendered and used to match against the target input image. Then a new pose is computed to increase the matching score. Existing methods for pose refinement use either hand-crafted image features~\citep{tjaden2017real} or matching score functions~\citep{rad2017bb8}. 

In this work, we propose \dimnet, a new refinement technique based on a deep neural network for iterative 6D pose matching. Given an initial 6D pose estimation of an object in a test image, \dimnet\ predicts a relative SE(3) transformation that matches a rendered view of the object against the observed image\yi{, or in other words, it predicts the relative rotation and translation that can refine the initial 6D pose estimation.}  By iteratively re-rendering the object based on the improved pose estimates, the two input images to the network become more and more similar, thereby enabling the network to generate more and more accurate pose estimates. Fig.~\ref{fig:intro} illustrates the iterative matching procedure of our network for pose refinement.

This work makes the following main contributions. i) We introduce a deep network for iterative, image-based pose refinement that does not require any hand-crafted image features and automatically learns an internal refinement mechanism. ii) We propose a disentangled representation of the SE(3) transformation between object poses to achieve accurate pose estimates. This representation also enables our approach to refine pose estimates of unseen objects. 
iii) We have conducted extensive experiments on the LINEMOD~\citep{hinterstoisser2012accv} and the Occlusion LINEMOD \citep{Brachmann2014Learning6O} datasets to evaluate the accuracy and various properties of \dimnet. These experiments show that our approach achieves large improvements over state-of-the-art RGB-only methods on both datasets. Furthermore, initial experiments demonstrate that \dimnet\ is able to accurately match poses for textureless objects (T-LESS~\citep{hodan2017t}) and for unseen objects \citep{wu20153d}. The rest of the paper is organized as follows. After reviewing related works in Section 2, we describe our approach for pose matching in Section 3. Experiments are presented in Section 4, and Section 5 concludes the paper.

\section{Related work}

We review representative works on 6D pose estimation in the literature.

\subsection{RGB based 6D Pose Estimation}

Traditionally, object pose estimation using RGB images is tackled by matching local features~\citep{lowe1999object,rothganger20063d,collet2011moped}. In this paradigm, a 3D model of an object is first reconstructed and local features of the object are attached to the 3D model. Keypoint-based features such as SIFT \citep{lowe1999object} or SURF \citep{bay2008speeded} are widely used. Given an input image, local features extracted from the image are matched against features on the 3D model. By filtering out incorrect matches using robust estimation techniques such as RANSAC \citep{nister2005preemptive}, the 6D pose of the object can be recovered using the 2D-to-3D correspondences between the local features. Local-feature matching based methods can handle partial occlusions between objects as long as the features on the visual part of the object are sufficient to determine the 6D pose. However, these methods cannot handle textureless objects well, since rich texture on the object is required in order to detect these features robustly.

In contrast, template-matching based methods are capable of handling textureless objects \citep{jurie2001real,liu2010fast,gu2010discriminative,hinterstoisser2012gradient}. In this paradigm, templates of an object are first constructed, where examples of templates are renderings of the object from the 3D object model or Histogram of Oriented Gradients (HOG) \citep{dalal2005histograms} templates from different viewpoints. Then these templates are matched against the input image to determine the location and orientation of the target object in the input image. The drawback of template-matching based methods is that they are not robust to occlusions between objects. When the target object is heavily occluded, the matching score is usually low which may result in incorrect pose estimation.

Recent approaches apply machine learning, especially deep learning, for 6D pose estimation using RGB images~\citep{Brachmann2014Learning6O,krull2015learning}. Learning techniques are employed to detect object keypoints for matching or learn better feature representations for pose estimation. The state-of-the-art methods~\citep{rad2017bb8,kehl2017ssd,tekin2017real,xiang2017posecnn,tremblay2018deep} augment deep learning based object detection or segmentation methods~\citep{girshick2015fast,long2015fully,liu2016ssd,redmon2016you} for 6D pose estimation. For example,~\citep{rad2017bb8,tjaden2017real,tremblay2018deep} utilize deep neural networks to detect keypoints on the objects, and then compute the 6D pose by solving the PnP problem. \citep{kehl2017ssd,xiang2017posecnn} employ deep neural networks to detect objects in the input image, and then classify or regress the detected object to its pose.  A recent work~\citep{sundermeyer2018implicit} uses an autoencoder to map the object in the image to a vector and search for the most similar vector in a pre-generated codebook for pose estimation. Overall, learning-based methods achieve better performance than traditional methods, largely due to the ability of learning a powerful feature representation for pose estimation.

\subsection{Depth based 6D Pose Estimation}

From another point of view, the 6D pose estimation problem can be tackled using depth images. Given a 3D model of an object and an input depth image, the problem is formulated as aligning the two point clouds computed from the 3D model and the depth image, respectively, which is also known as the geometric registration problem. Roughly speaking, geometric registration methods can be classified as local refinement methods and global registration methods. The most well-known local refinement algorithm is the Iterative Closest Point (ICP) algorithm \citep{besl1992method} and its variants \citep{rusinkiewicz2001efficient,salvi2007review,tam2013registration}. Given an initial pose estimation, the ICP algorithm iterates between finding the correspondences between points and refining the pose estimation using the new correspondences. In general, local refinement algorithms are sensitive to the initial pose. If the initial pose estimation is not close enough, the algorithm may converge to a local mimimum.

Global registration methods \citep{mellado2014super,theiler2015globally,zhou2016fast,yang2016go} solve a more challenging problem by not assuming an initial pose estimate. A common strategy is to utilize iterative model fitting frameworks such as RANSAC. In each iteration, a set of point correspondences are sampled, and an alignment is computed and evaluated using the sampled correspondences. The limitation of most global registration methods is that they are computationally expensive. Also, the registration quality heavily depends on the quality of the 3D model and the scanned point cloud. In order to improve the registration performance, features on point clouds are also introduced for matching. These include point pairs \citep{mian2006three,hinterstoisser2016going}, spin-images \citep{johnson1999using}, and point-pair histograms \citep{rusu2009fast,tombari2010unique}. Similar to the trend in image-based matching, recent approaches \citep{wang2019densefusion} propose to learn point features for registration, such as applying deep neural networks to point clouds \citep{qi2017pointnet}.

\subsection{RGB-D based 6D Pose Estimation} 

When both RGB images and depth images are available, they can be combined to improve 6D pose estimation. A common strategy is to estimate an initial pose of an object based on the color image, and then refine the pose using depth-based local refinement algorithms such as ICP~\citep{hinterstoisser2012accv,michel2016global,zeng2017multi}.

For example,~\cite{hinterstoisser2012accv} renders the 3D model of an object into templates of color images, and then matches these templates against the input image to estimate an initial pose. The final pose estimation is obtained via ICP refinement on the initial pose. ~\cite{Brachmann2014Learning6O},~\cite{brachmann2016uncertainty},~\cite{michel2016global} regress each pixel on the object in the input image to the 3D coordinate of that pixel on the 3D model. When depth images are available, the 3D coordinate regression establishes correspondences between 3D scene points and 3D model points, from which the 6D pose can be computed by solving a least-squares problem. PoseCNN \citep{xiang2017posecnn} introduces an end-to-end neural network for 6D object pose estimation using RGB images only. Given an initial pose from the network, a customized ICP method is applied to refine the pose.  A recent work \citep{wang2019densefusion} introduces a neural network that combines RGB images and depth images for 6D pose estimation, and an iterative pose refinement network using point clouds as input.

\subsection{RGB vs. RGB-D}

Overall, the performance of RGB-based methods is still not comparable to that of the RGB-D based methods. We believe that this performance gap is largely due to the lack of an effective pose refinement procedure using RGB images only. \cite{manhardt2018deep} which is published at the same time as ours introduces a method to refine 6D object poses with only RGB images, but there is still a large performance gap between \cite{manhardt2018deep} and depth-based methods. Our work is complementary to existing 6D pose estimation methods by providing a novel iterative pose matching network for pose refinement on RGB images.

The approaches most related to ours are the object pose refinement network in~\cite{rad2017bb8} and the iterative hand pose estimation approaches in~\cite{carreiran2016hum,oberweger2015tra}. Compared to these techniques, our network is designed to directly regress to relative SE(3) transformations. We are able to do this due to our disentangled representation of rotation and translation and the reference frame we used for rotation, which also allows our approach to match unseen objects. As shown in \cite{mousavian20173d}, the choice of reference frame is important to achieve good pose estimation results. Our work is also related to recent visual servoing methods based on deep neural networks \citep{saxena2017exploring,costante2018ls} that estimate the relative camera pose between two image frames, while we focus on 6D pose refinement of objects. Recent works \citep{garon2016real,garon2017deep} that focus on tracking could predict the transformation of the object pose between previous frame and current frame and have the potential to be used for pose refinement.

%\TODO{related visual servoing works}
%\ddd{Need to cite Mousavian's paper on 3d boxes and their discussion of pose representation. See paper link in latex file.}
%http://openaccess.thecvf.com/content_cvpr_2017/papers/Mousavian_3D_Bounding_Box_CVPR_2017_paper.pdf

%-------------------------------------------------------------------
%-------------------------------------------------------------------
\section{\dimnet\ Framework}
%-------------------------------------------------------------------
%-------------------------------------------------------------------

%\old{use \textbackslash old\{\} to comment old sentence} \\
%\new{use \textbackslash new\{\} to highlight new sentence} \\
%\TODO{use \textbackslash TODO\{\} to say TODO things}

%-------------------------------------------------------------------

In this section, we describe our deep iterative matching network for 6D pose estimation. Given an \real\ image and an initial pose estimate of an object in the image, we design the network to directly output a relative SE(3) transformation that can be applied to the initial pose to improve the estimate. We first present our strategy of zooming in the \real\ image and the \rend\ image that are used as inputs of the network. Then we describe our network architecture for pose matching. After that, we introduce a disentangled representation of the relative SE(3) transformation and a new loss function for pose regression. Finally, we describe our procedure for training and testing the network.

\subsection{High-resolution Zoom In}
%-------------------------------------------------------------------

\label{sec:zoom}

\begin{figure*}[t] 
	\centering
	\includegraphics[width =.8\linewidth]{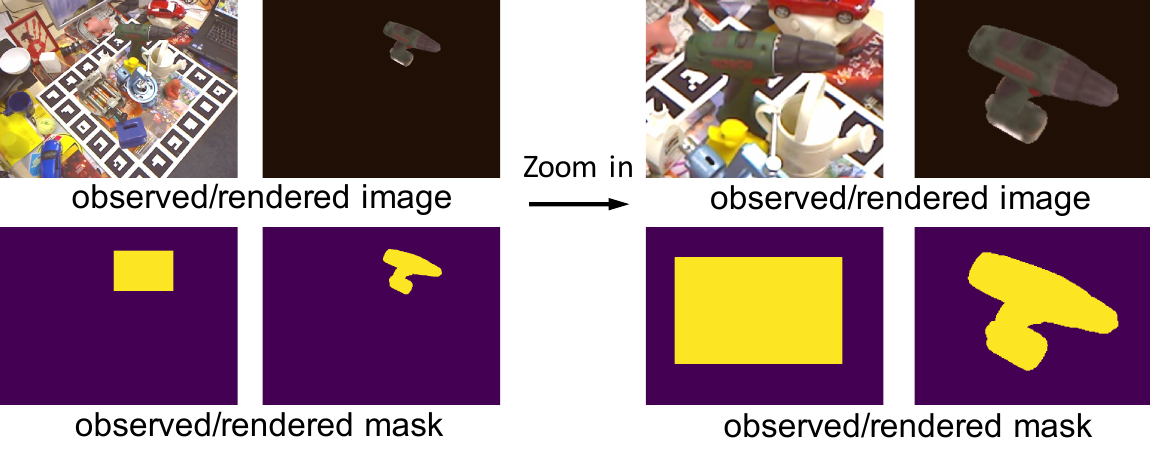}
	\caption{\dimnet\ operates on a zoomed in, up-sampled input image, the rendered image, and the two object masks ($480\times640$ in our case after zooming in). \yi{More specifically, we enlarge the bounding box of the object in the rendered image, crop the corresponding patch using the enlarged bounding box in both image pairs and mask pairs and then up-sample them to high resolution. Notice that the aspect ratio is kept during this process to avoid image distortion. See Sec.~\ref{sec:zoom} for more details.}}
	\label{fig:zoom}
\end{figure*}

It can be difficult to extract useful features for matching if objects in the input image are very small. 
To obtain enough details for pose matching, we zoom in the \real\ image and the \rend\ image before feeding them into the network, as shown in Fig.~\ref{fig:zoom}. 
Specifically, in the $i$-th stage of the iterative matching, given a 6D pose estimate $\mathbf{p}^{(i-1)}$ from the previous step, we render a synthetic image using the 3D object model viewed according to $\mathbf{p}^{(i-1)}$. 

We additionally generate one foreground mask for the \real\ image and \rend\ image. The four images are cropped using an enlarged bounding box according to the \real\ mask and the \rend\ mask, where we make sure the enlarged bounding box has the same aspect ratio as the input image and is centered at the 2D projection of the origin of the 3D object model. 

In more detail, given the \rend\ mask $\mrend$ and the \real\ mask $\mreal$, the cropping patch is computed as
\begin{equation}
\begin{split}
x_{\text{dist}} = \max(&|l_{\text{obs}}-x_{c}|, |l_{\text{rend}}-x_{c}|,\\
			  &|r_{\text{obs}}-x_{c}|, |r_{\text{rend}}-x_{c}|), \\
y_{\text{dist}} = \max(&|u_{\text{obs}}-y_{c}|, |u_{\text{rend}}-y_{c}|,\\
			  &|d_{\text{obs}}-y_{c}|, |d_{\text{rend}}-y_{c}|), \\
\text{width} = \max(&x_{\text{dist}}, y_{\text{dist}} \cdot r) \cdot 2\lambda, \\
\text{height} = \max(&x_{\text{dist}} / r, y_{\text{dist}}) \cdot 2\lambda,
\end{split}
\label{eq.zoom_bound}
\end{equation}
where $u_{*}, d_{*}, l_{*}, r_{*}$ denotes the upper, lower, left, right bound of foreground mask of observed or rendered images, $x_{c}, y_{c}$ represent the 2D projection of the center of the object in $\xrend$, $r$ represent the aspect ratio of the origin image (width/height), $\lambda$ denotes the expand ratio, which is fixed to 1.4 in the experiment in order to make the expanded patch is roughly twice than the nested one. Then this patch is bilinearly sampled to the size of the original image, which is $480 \times 640$ in this paper. By doing so, not only the object is zoomed in without being distorted, but also the network is provided with the information about where the center of the object lies.

%-------------------------------------------------------------------
\subsection{Network Structure}
%-------------------------------------------------------------------
\label{sec.network_structure}
\begin{figure*}[t]
	\centering
	\includegraphics[width =\linewidth]{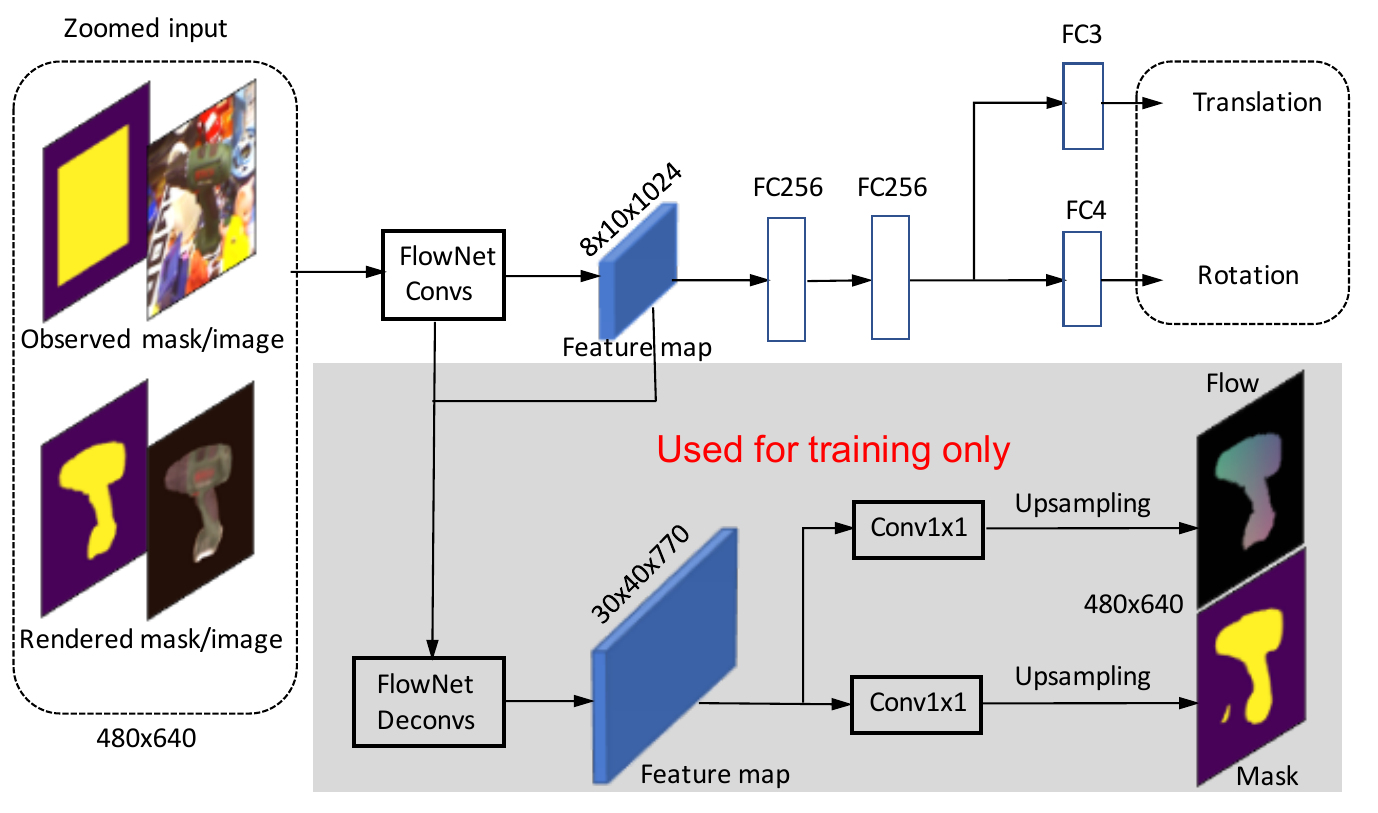}
	\caption{{\dimnet\ uses a FlowNetSimple backbone to predict a relative SE(3) transformation to match the observed and \rend\ image of an object. \yi{Taking observed image and rendered image and their corresponding masks as input, the convolution layers output a feature map which then be forwarded through several fully connected layers to predict the translation and rotation. The same feature map, combined with feature maps in the previous layers, will also be used to predict flow and foreground mask during training.}}}
	\label{fig:network}
    
\end{figure*}

Fig.~\ref{fig:network} illustrates the network architecture of \dimnet. 
The observed image, the \rend\ image, and the two masks, are concatenated into an eight-channel tensor input to the network (3 channels for \real/\rend\ image, 1 channel for each mask). 
We use the FlowNetSimple architecture from~\cite{dosovitskiy2015flownet} as the backbone network, which is trained to predict optical flow between two images.
We tried using the VGG16 image classification network~\citep{simonyan2014very} as the backbone network, but the results were very poor, confirming the intuition that a representation related to optical flow is very useful for pose matching\yi{~\citep{wang2017deepvo}}.

The pose estimation branch takes the feature map after 10 convolution layers from FlowNetSimple as input. It contains two fully-connected layers each with dimension 256, followed by two additional fully-connected layers for predicting the quaternion of the 3D rotation and the 3D translation, respectively. 

During training, we also add two auxiliary branches to regularize the feature representation of the network and increase training stability \yi{and performance, see Sec.~\ref{sec.exp_on_LM} and Table. \ref{table.ablation_to_mask_and_flow} for more details}. One branch is trained for predicting optical flow between the \rend\ image and the \real\ image, and the other branch for predicting the foreground mask of the object in the \real\ image. 

%Compared with FlowNetCorr, FlowNetSimple takes the advantage of fast speed, keeping feature from both real image and reference image, and good at capture small displacement between real image and reference image.

%%-------------------------------------------------------------------
\subsection{Disentangled Transformation Representation}
\label{sec:untangle}
%-------------------------------------------------------------------

\begin{figure*}[t]
%\subfloat[Init State\label{subfig.init_coord_a}]{
\includegraphics[width=0.23\textwidth]{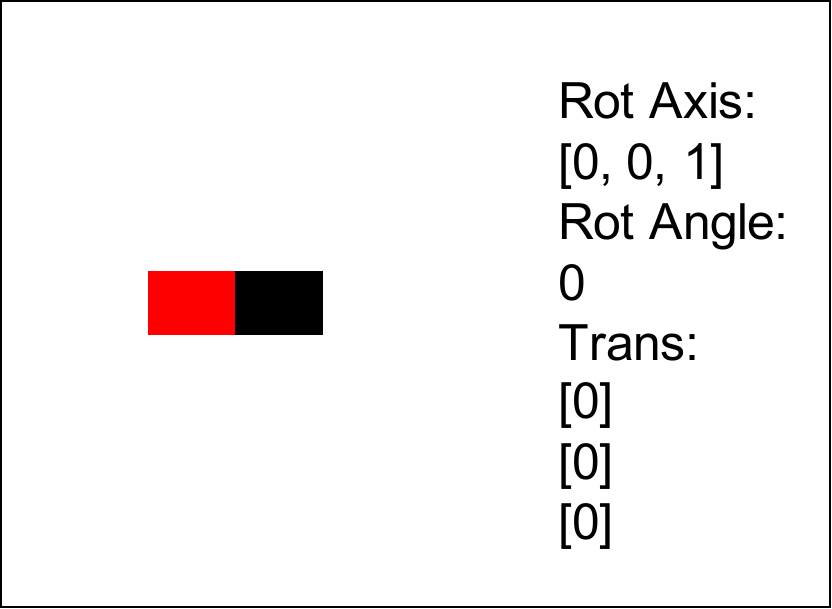}%}
\hfill
%\subfloat[Naive Coordinate\label{subig.naive_coord_a}]{
\includegraphics[width=0.23\textwidth]{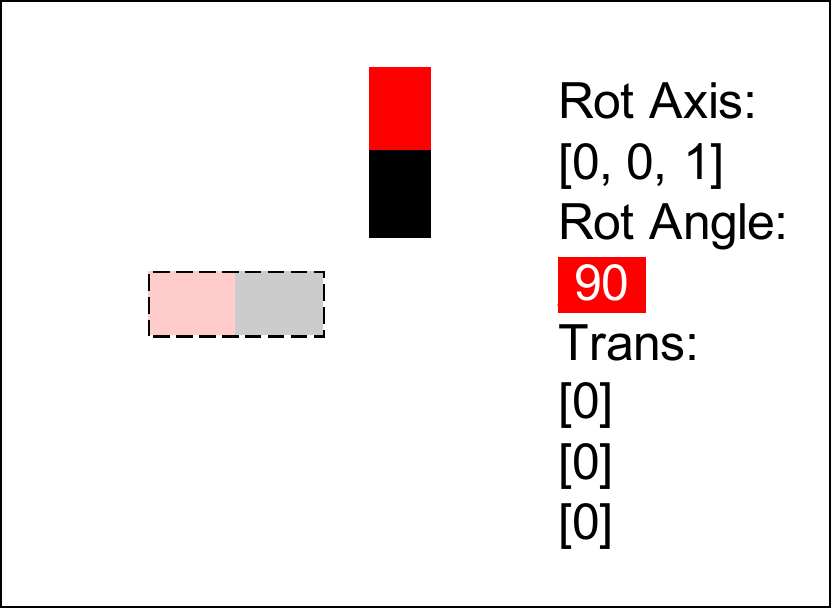}%}
\hfill
%\subfloat[Model Coordinate\label{subfig.model_coord_a}]{
\includegraphics[width=0.23\textwidth]{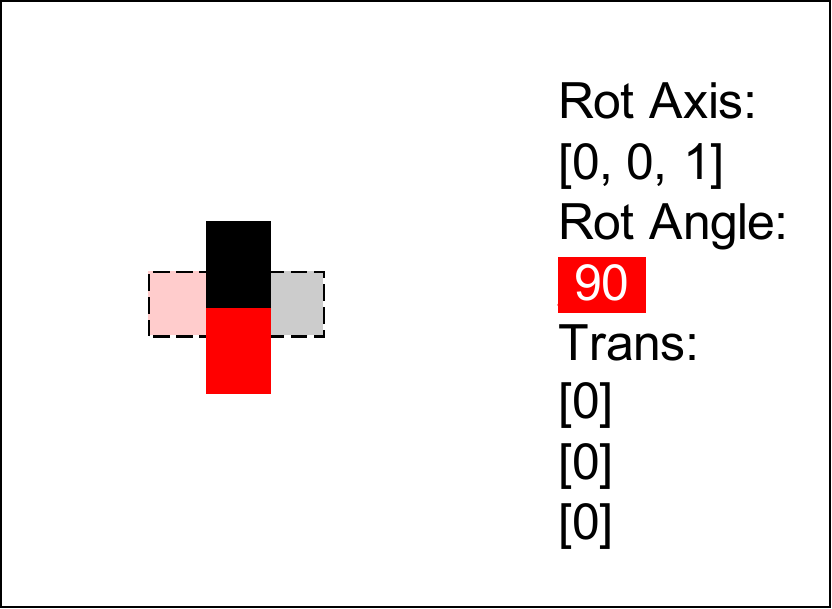}%}
\hfill
%\subfloat[Our Coordinate\label{subfig.camera_coord_a}]{
\includegraphics[width=0.23\textwidth]{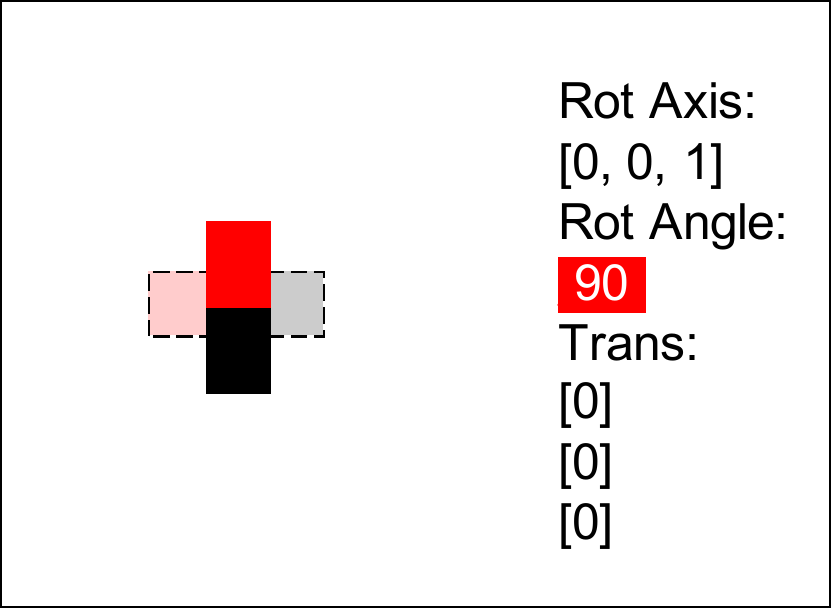}%}
%\caption{\small{Rotation results using same rotation vector and different coordinate systems. Noticing that rotation under naive coordinate will leads to not only a rotation but also a translation. For the model coordinate, as the frame of object model can be defined arbitrary, one object can rotate along any axis given the same rotation vector.}}
% \label{fig.different_reprentation_a}
%\end{figure*}

%\begin{figure*}[t]
\subfloat[Initial pose\label{subfig.init_coord_b}]{
\includegraphics[width=0.23\textwidth]{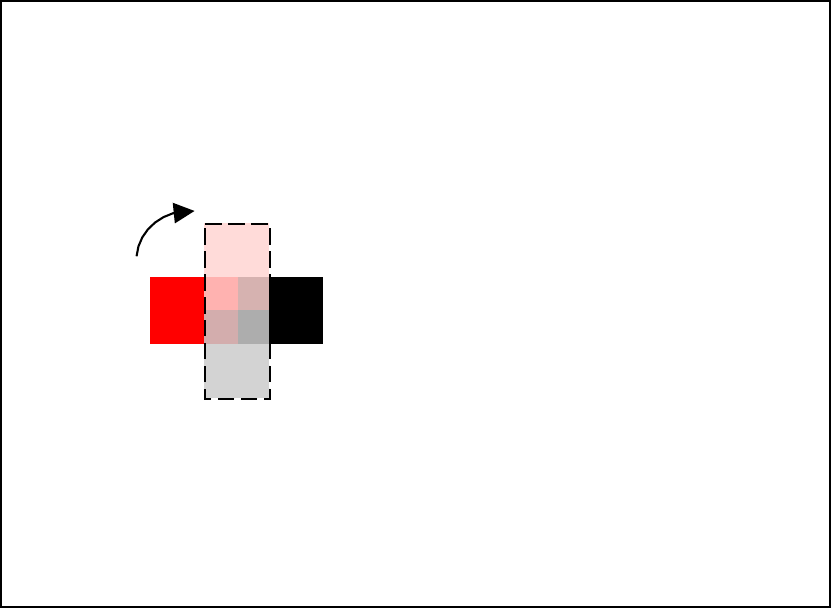}}
\hfill
\subfloat[Camera coord.\label{subig.naive_coord_b}]{
\includegraphics[width=0.23\textwidth]{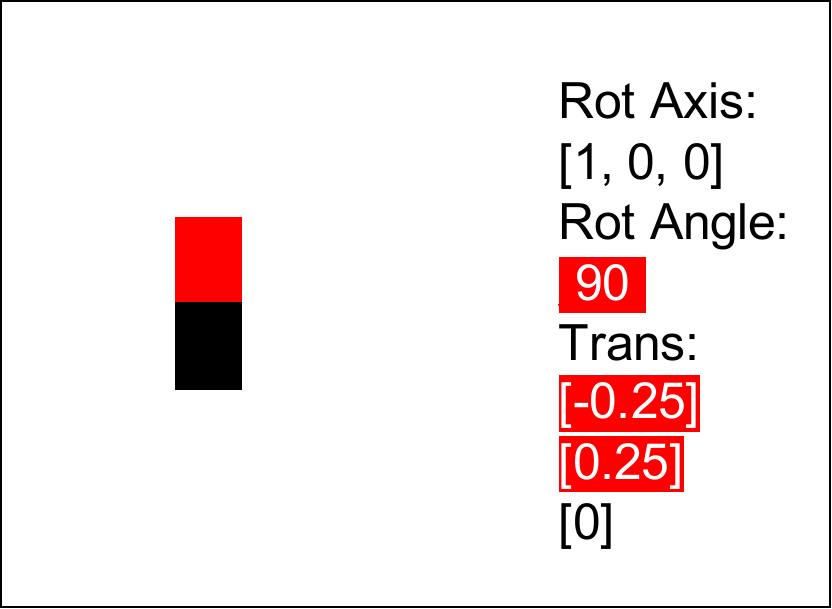}}
\hfill
\subfloat[Model coord.\label{subfig.model_coord_b}]{
\includegraphics[width=0.23\textwidth]{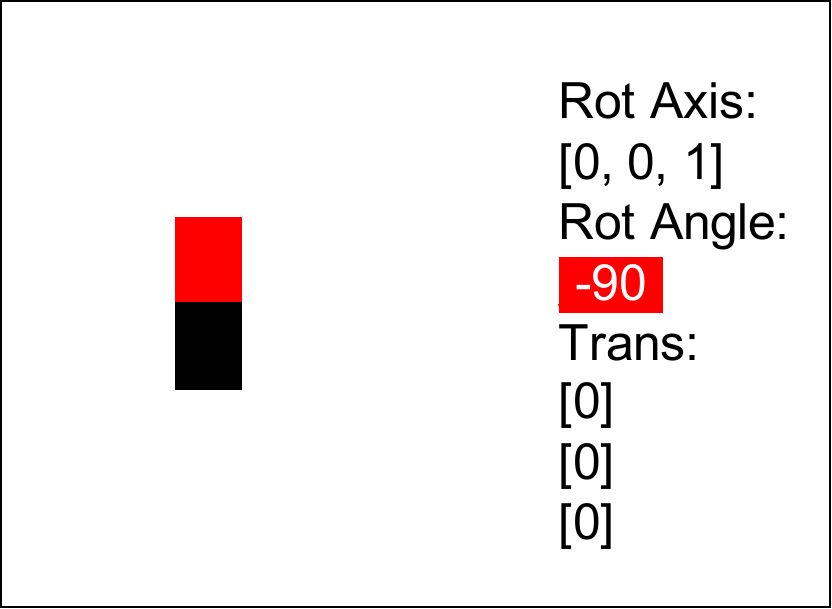}}
\hfill
\subfloat[\small{disentangled coord.}\label{subfig.camera_coord_b}]{
\includegraphics[width=0.23\textwidth]{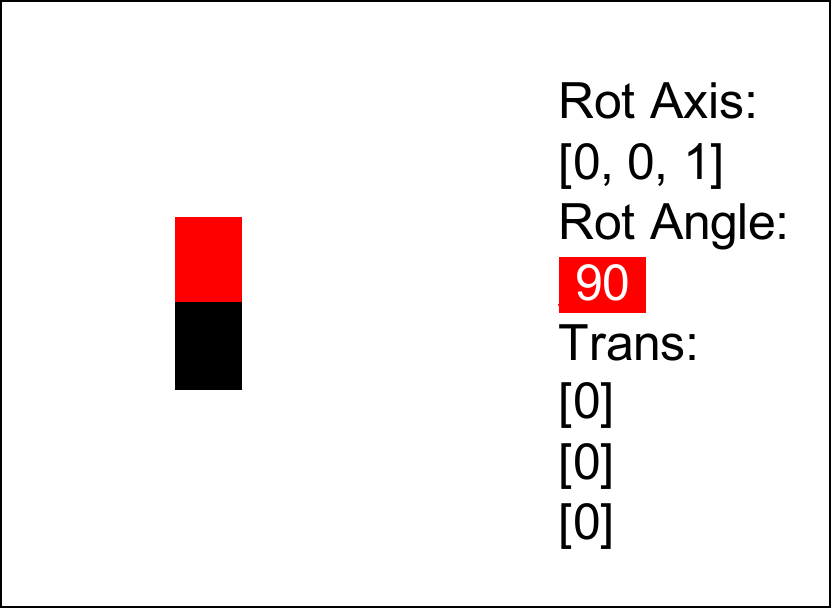}}\vspace*{-2ex}
\caption{\small{Rotations using different coordinate systems. (Upper row) The panels show how a 90 degree rotation in the image plane axis changes the position of the object shown in column (a). In the camera coordinate system, the center of rotation is in the center of the image, thereby causing an undesired  translation in addition to the object rotation. In the model coordinate frame, as the frame of the object model can be defined arbitrarily, an object might rotate along any axis given the same rotation vector.  Shown here is a CCW rotation, but the same axis might also result in an out of plane rotation for a differently defined object coordinate frame.  In our disentangled representation, the center of rotation is in the center of the object and the axes are defined parallel to the camera axes. As a result, a rotation around a specific axis always results in the same object rotation, independent of the object. (Lower row) Rotation vectors a network would have to predict in order to achieve an in-place rotation using the different coordinate systems. Notice the extra translations required to compensate for the translation caused by the rotation using camera coordinates (column b). In model coordinates, the network would have to learn the frame specified for the object model in order to determine the correct rotation axis and angle.  In our disentangled representation, rotation axis and angle are independent of the object.}}
\label{fig.rot_reprentations}\vspace*{-1ex}
%\end{figure*}

%\begin{figure*}[t]
\subfloat[Camera coord. xy-plane translation\label{subfig.naive_trans_xy}]{
\includegraphics[width=0.23\textwidth]{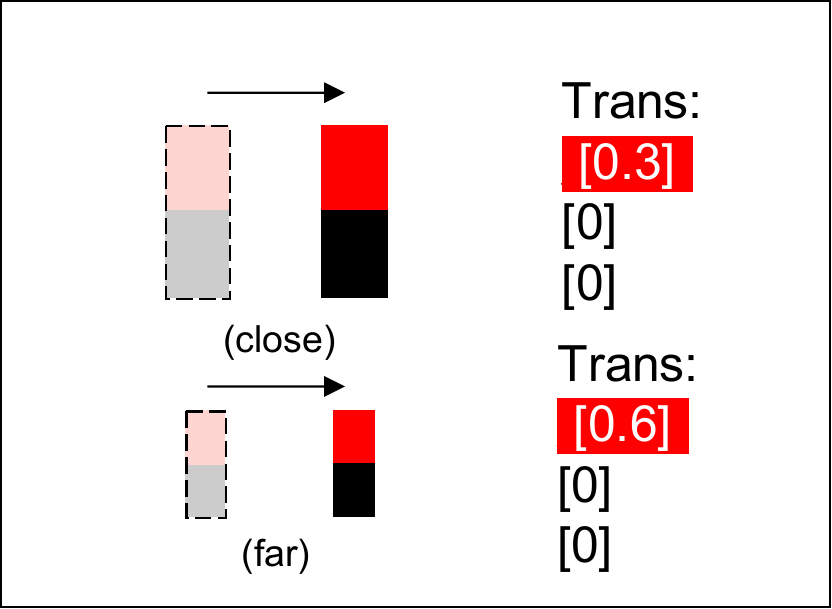}}
\hfill
\subfloat[Disentangled coord. xy-plane translation\label{subig.camera_trans_xy}]{
\includegraphics[width=0.23\textwidth]{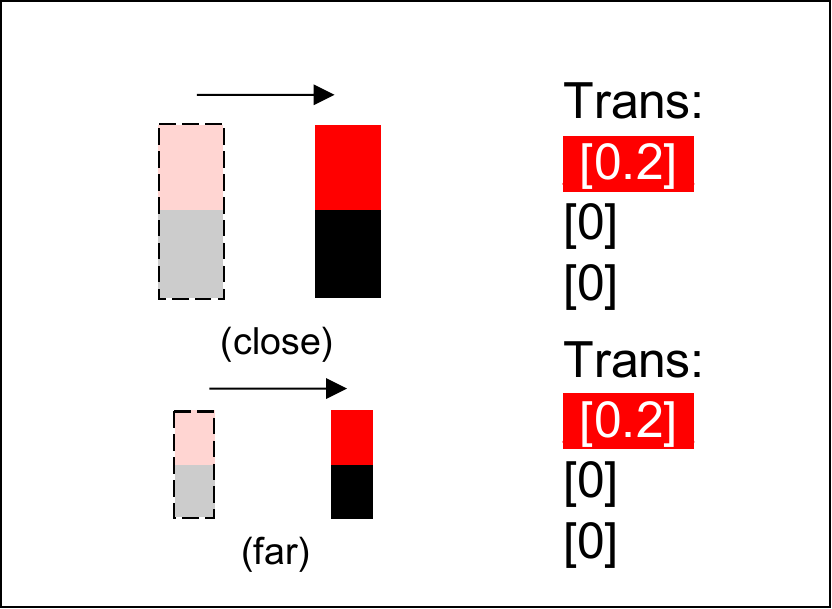}}
\hfill
\subfloat[Camera coord. z-axis translation\label{subfig.naive_trans_z}]{
\includegraphics[width=0.23\textwidth]{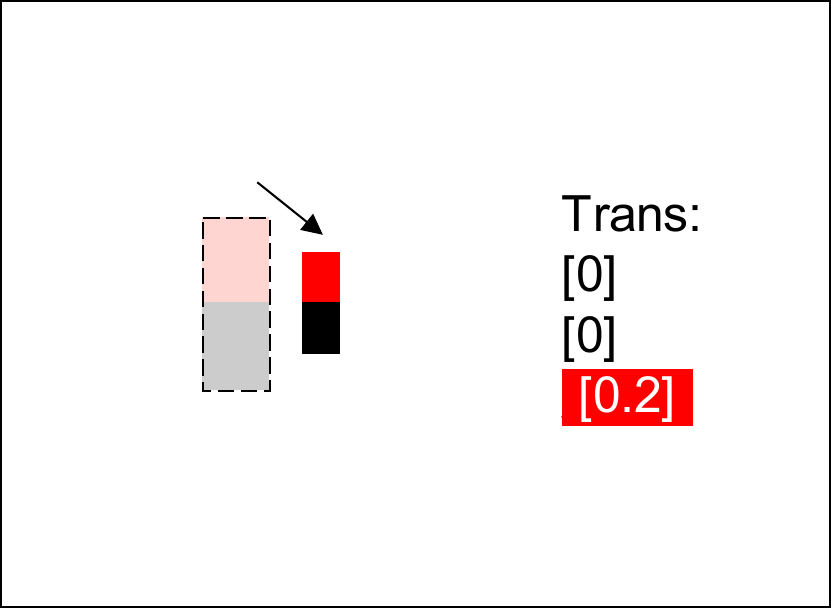}}
\hfill
\subfloat[Disentangled coord. z-axis translation\label{subfig.camera_trans_z}]{
\includegraphics[width=0.23\textwidth]{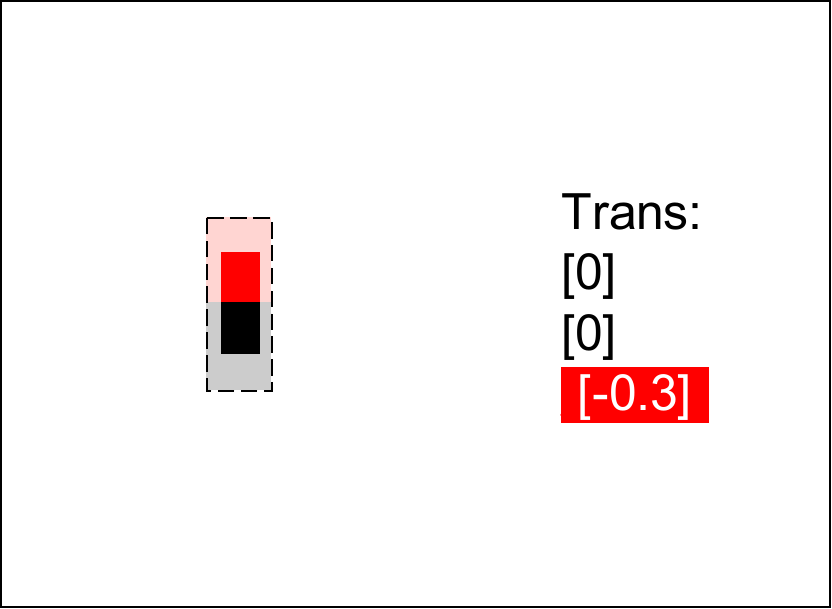}}\vspace*{-2ex}
\caption{\small{Translations using camera and our disentangled representations. In camera coordinates,  translations in the image plane are represented by vectors in 3D space.  As a result, the same translation in the 2D image corresponds to different translation vectors depending on whether an object is close or far from the camera. In our disentangled representation, the value of x and y is only related to the 2D vector in the image-plane. Additionally, as shown in column (c), in the camera representation, a translation along the z-axis is not only difficult to infer from the image, but also causes a move relative to the center of the image. In our disentangled translation representation (column (d)), only the change of scale needs to be estimated, making it independent of other translations and the metric size and distance of the object.}}
\label{fig.trans_reprentations}\vspace*{-3ex}
\end{figure*}

%In a nutshell, we want the following properties from this representation. First, it should be independent of the specific local coordinate frame specified for a 3D object model. Otherwise, the network would have to learn this coordinate frame for every object, potentially reducing its performance and making it inapplicable to unknown objects. Second, the pose changes predicted by the network should be as closely as possible related to consistent operations in the input images. We achieve this by choosing coordinate frames such that 3D object translations correspond to pixel-wise translations in the image plane along with a change in scale, and 3D rotations correspond to rotations around the image center along with a pan and tilt of the image. Using such a representation, the \dimnet\ network  can operate independently of the actual size of the object and its internal model coordinate framework. It only has to learn to transform the \rend\ image such that it becomes more similar to the observed image.

%using a quaternion or Euler angle for the rotation matrix $\mathbf{R}_\Delta$ and a vector for the translation $\mathbf{t}_\Delta$, so that the full transformation matrix can be written as $[\mathbf{R_\Delta}|\mathbf{t}_\Delta]$

The representation of the coordinate frames and the relative SE(3) transformation $\mathbf{\Delta p}$ between the current pose estimate and the target pose has important ramifications for the performance of the network.  Ideally, we would like (1) the individual components of these transformations to be maximally dis-entangled, thereby not requiring the network to learn unnecessarily complex geometric relationships between translations and rotations, and (2) the transformations to be independent of the intrinsic camera parameters and the actual size and coordinate system of an object, thereby enabling the network to reason about changes in object appearance rather than accurate distance estimates. 

The most obvious choice are camera coordinates to represent object poses and transformations.  Denote the relative rotation and translation as $[\mathbf{R_\Delta}|\mathbf{t}_\Delta]$ \yi{(We denote $\mathbf{R}_*$ as rotation and and $\mathbf{t}_*$ as translation in this paper)}.  Given a source object pose $[\mathbf{R}_{\text{src}}|\mathbf{t}_{\text{src}}]$, the transformed target pose would be as follows:
\begin{equation}
\begin{split}
\mathbf{R}_\text{tgt} = \mathbf{R}_\mathrm{\Delta} \mathbf{R}_\text{src}, \quad
\mathbf{t}_\text{tgt} = \mathbf{R}_\mathrm{\Delta} \mathbf{t}_\text{src}+\mathbf{t}_\mathrm{\Delta},
\end{split}
\label{eq.naive_transform}
\end{equation}
where $[\mathbf{R}_{\text{tgt}}|\mathbf{t}_{\text{tgt}}]$ denotes the target pose resulting from the transformation.
The $\mathbf{R}_\mathrm{\Delta} \mathbf{t}_\text{src}$ term indicates that a rotation will cause the object not only to rotate, but also translate in the image even if the translation vector $\mathbf{t}_\mathrm{\Delta}$ equals to zero. 
Column (b) in Fig.~\ref{fig.rot_reprentations} illustrates this connection for an object rotating in the image plane.  
In standard camera coordinates, the translation $\mathbf{t}_\mathrm{\Delta}$ of an object is in the 3D metric space (meter, for instance), which couples object size with distance in the metric space.  This would require the network to memorize the actual size of each object in order to transform mis-matches in images to distance offsets. It is obvious that such a representation is not appropriate, particularly for matching unknown objects. 
\if 0
\begin{figure*}[t]
	\centering
	\includegraphics[width = \linewidth]{4_rotation_2d_v4.pdf}
\caption{Three different coordinate systems for the relative rotation.}% All the coordinate system follows right-handed Cartesian coordinates.}
\label{fig.different_reprentation}
\end{figure*}
\fi

To eliminate these problems, we propose to decouple the estimation of $\mathbf{R}_\mathrm{\Delta}$ and $\mathbf{t}_\mathrm{\Delta}$. First, we move the center of rotation from the origin of the camera to the center of the object in the camera frame, given by the current pose estimate. In this representation, a rotation does not change the translation of the object in the camera frame. The remaining question is how to choose the directions of the rotational axes of the coordinate frame. One way is to use the axes as specified in the 3D object model. However, as illustrated in column (c) of Fig.~\ref{fig.rot_reprentations}, such a representation would require the network to learn and memorize the coordinate frames of each object, which makes  training more difficult and cannot be generalized to pose matching of unseen objects.  Thus, we propose to use axes parallel to the axes of the camera frame when computing the relative rotation. By doing so, the network can be trained to estimate the relative rotation independently of the coordinate frame of the 3D object model, as illustrated in column (d) in Fig.~\ref{fig.rot_reprentations}. 

%Our experiments in Sec.~\ref{sec.exp_on_LM} also show that such a representation produces far better results than using the axes of the coordinate frame of the 3D object model.

%rotations of the object position always correspond to the same operations in the \rend\ image. Specifically, rotations around individual axes approximately result in image rotations and changes in aspect ratios along the image dimensions (plus changes in visible parts of the object).

In order to estimate the relative translation, let $\mathbf{t}_{\text{tgt}}=(x_{\text{tgt}}, y_{\text{tgt}}, z_{\text{tgt}})$ and $\mathbf{t}_{\text{src}}=(x_{\text{src}}, y_{\text{src}}, z_{\text{src}})$ be the target translation and the source translation. A straightforward way to represent translation is  $\mathbf{t}_\mathrm{\Delta} = (\mathrm{\Delta}_x, \mathrm{\Delta}_y, \mathrm{\Delta}_z) = \mathbf{t}_{\text{tgt}} - \mathbf{t}_{\text{src}} $. However, it is not easy for the network to estimate the relative translation in the 3D metric space given only 2D images without depth information. The network has to recognize the size of the object, and map the translation in 2D space to 3D according to the object size. Such a representation is not only difficult for the network to learn, but also has problems when dealing with unknown objects or objects with similar appearance but different sizes. Instead of training the network to directly regress to the vector in the 3D space, we propose to regress to object changes in the 2D image space. Specifically, we train the network to regress to the relative translation $\mathbf{t}_\mathrm{\Delta} = (v_x, v_y, v_z)$, where $v_x$ and $v_y$ denote the number of pixels the object should move along the image x-axis and y-axis and $v_z$ is  the scale change of the object:
\begin{equation}
\begin{split}
v_x &= f_x (x_{\text{tgt}} / z_{\text{tgt}} - x_{\text{src}} / z_{\text{src}}), \\
v_y &= f_y (y_{\text{tgt}} / z_{\text{tgt}} - y_{\text{src}} / z_{\text{src}}), \\
v_z &= \log(z_{\text{src}} / z_{\text{tgt}}),
\end{split}
\label{eq.2d_translation}
\end{equation}
where $f_x$ and $f_y$ denote the focal lengths of the camera. The scale change $v_z$ is defined to be independent of the absolute object size or distance by using the ratio between the distances of the rendered and observed object. We use logarithm for $v_z$ to make sure that a value of zero corresponds to no change in scale or distance. Considering the fact that $f_x$ and $f_y$ are constant for a specific dataset, we simply fix it to 1 in training and testing the network.

%The last line in Eq.~\ref{eq.2d_translation} recovers the updated distance from $v_z$.

%Furthermore, by moving the center of rotation into the object, the translation is fully decoupled from the rotation: 
%\begin{equation}
%\begin{split}
%x_{tgt} &= z_{tgt} \cdot (v_x + x_{src} / z_{src})\\
%y_{tgt} &= z_{tgt} \cdot (v_y + y_{src} / z_{src})\\
%z_{tgt} &= z_{src} / \exp(v_z)
%x_{tgt} &= z_{src} v_x + x_{src} \\
%y_{tgt} &= z_{src} v_y + y_{src} \\
%z_{tgt} &= z_{src} / \exp(v_z)
%\end{split}
%\label{eq.2d_translation}
%\end{equation}
%Here, $\mathbf{t}_{tgt}=(x_{tgt}, y_{tgt}, z_{tgt})$, $\mathbf{t}_{src}=(x_{src}, y_{src}, z_{src})$ are object translations in the camera frame, and $v_x$, $v_y$, and $v_z$ are the target outputs for the network, representing  movement along the image x-axis and y-axis, and scale changes, respectively. The scale value $v_z$, is chosen to be independent of the absolute object size or distance; it is defined via the ratio between the distances of the rendered and observed object. We use $\log{\frac{z_{src}}{z_{tgt}}}$ for $v_z$ to make sure that 0 corresponds to no change in scale, or distance.  The last line in Eq.~\ref{eq.2d_translation} recovers the updated distance from $v_z$.

Our representation of the relative transformation has several advantages. First, rotation does not influence the estimation of translation, so that the translation no longer needs to offset the movement caused by rotation around the camera center. Second, the intermediate variables $v_x$, $v_y$, $v_z$ represent simple translations and scale change in the image space. Third, this representation does not require any prior knowledge of the object. Using such a representation, the \dimnet\ network  can operate independently of the actual size of the object, its internal model coordinate framework, and the camera intrinsics. It only has to learn to transform the \rend\ image such that it becomes more similar to the observed image.

%\begin{equation}
%\mathbf{R_{tgt}=R_{src}R_\delta}
%\label{eq.model_coordinate}
%\end{equation}

%-------------------------------------------------------------------
\subsection{Matching Loss}
\label{sec.points_matching_loss}
%-------------------------------------------------------------------

A straightforward way to train the pose estimation network is to use separate loss functions for rotation and translation. For example, we can use the angular distance between two rotations to measure the rotation error and use the $\ell_2$ distance to measure the translation error. However, using two different loss functions for rotation and translation suffers from the difficulty of balancing the two losses. \citep{kendall2017geometric} proposed a geometric reprojection error as the loss function for pose regression that computes the average distance between the 2D projections of 3D points in the scene using the ground truth pose and the estimated pose.
%as Eq.`\ref{eq.reproj2d}.
% \begin{equation}
% ReProj(\mathbf{p}, \mathbf{\hat{p}}) = \frac{1}{n} \sum_{j=1}^n L_1(2D\_proj(\mathbf{x}_j, \mathbf{p}) - 2D\_proj(\mathbf{x}_j, \mathbf{\hat{p}}))
% \label{eq.reproj2d}
% \end{equation}
Considering the fact that we want to accurately predict the object pose in 3D, we introduce a modified version of the geometric reprojection loss in \citep{kendall2017geometric}, and we call it the Point Matching Loss. Given the ground truth pose $\mathbf{p} = [\mathbf{R}|\mathbf{t}]$ and the estimated pose $\mathbf{\hat{p}}=[\mathbf{\hat{R}}|\mathbf{\hat{t}}]$, the point matching loss is computed as:
\begin{equation}
L_\text{pose}(\mathbf{p}, \mathbf{\hat{p}})
= \frac{1}{n} \sum_{i=1}^n \|(\mathbf{R}\mathbf{x}_i+ \mathbf{t}) - (\mathbf{\hat{R}}
\mathbf{x}_i+\mathbf{\hat{t}})\|_1,
\label{eq.reproj3d}
\end{equation}
where $\mathbf{x}_i$ denotes a randomly selected 3D point on the object model and $n$ is the total number of points (we choose 3,000 points in our experiments). \yi{The formulation of point matching loss is similar to the one used to compute average distance (ADD) metric in Eq. \ref{eq.add}. The main difference is that other than using $\ell_2$ norm, point matching loss computes the average $\ell_1$ distance between 3D points transformed by the ground truth pose and the estimated pose in order to avoid the large graident caused by outliers and maintain the stability of loss during training}. In this way, it measures how the transformed 3D models match against each other for pose estimation. \citep{xiang2017posecnn} also uses a variant of the point matching loss for rotation regression.

%and $\theta$ is used to normalize the loss since models in different datasets may have very different scales (we fix it to 0.1 which represents 10 centimeters in this paper).

%\ddd{why do you need scaling if you apply the same value for $\theta$ to the whole dataset? Why not a $\theta$ for each object?}
%\yi{I didn't realize that $\theta$ could be different for different objects}

%-------------------------------------------------------------------
\subsection{Training and Testing}
%-------------------------------------------------------------------

In training, we assume that we have 3D object models and images annotated with ground truth 6D object poses. By adding noises to the ground truth poses as the initial poses, we can generate the required \real\ and \rend\ inputs to the network along with the pose target output that is the pose difference between the ground truth pose and the noisy pose. Then we can train the network to predict the relative transformation between the initial pose and the target pose.

During testing, we find that the iterative pose refinement can significantly improve the accuracy. To see, let $\pose^{(i)}$ be the pose estimate after the $i$-th iteration of the network. If the initial pose estimate $\pose^{(0)}$ is relatively far from the correct pose, the rendered image $\xrend(\pose^{(0)})$ may have only little viewpoint overlap with the observed image $\xreal$. In such cases, it is very difficult to accurately estimate the relative pose transformation $\mathbf{\Delta p}^{(0)}$ directly. This task is even harder if the network has no priori knowledge about the object to be matched.  
In general, it is reasonable to assume that if the network improves the pose estimate $\mathbf{p}^{(i+1)}$ by updating 
$\mathbf{p}^{(i)}$ with $\mathbf{\Delta p}^{(i)}$ in the $i$-th iteration, then the image rendered according to this new estimate, $\xrend(\mathbf{p}^{(i+1)})$ is also more similar to the \real\ image $\xreal$ than $\xrend(\mathbf{p}^{(i)})$ was in the previous iteration, thereby providing input that can be matched more accurately.

However, we found that, if we train the network to regress the relative pose in a single step, the estimates of the trained network do not improve over multiple iterations in testing.
To generate a more realistic data distribution for training similar to testing, we perform multiple iterations during training as well. Specifically, for each training image and pose, we apply the transformation predicted from the network to the pose and use the transformed pose estimate as another training example for the network in the next iteration. By repeating this process multiple times, the training data better represents the test distribution and the trained network also achieves significantly better results during iterative testing (such an approach has also proven useful for iterative hand pose matching~\citep{oberweger2015tra} \yi{and image alignment~\citep{lin2017inverse}}).

\section{Experiments}

We conduct extensive experiments on the LINEMOD dataset \citep{hinterstoisser2012accv} and the Occlusion LINEMOD dataset \citep{Brachmann2014Learning6O} to evaluate our \dimnet\ framework for 6D object pose estimation. We test different properties of \dimnet\ and show that it surpasses other RGB-only methods by a large margin. We also show that our network can be applied to pose matching of unseen objects during training.

\subsection{Training Implementation Details}
\label{sec.training_details}
%Here we introduce some general settings used in training and testing.

% \begin{equation}
% L_{flow} = \frac{1}{n} \sum_{i=1}^n L_2(flow_{gt}(\mathbf{x}_j)-flow_{est}(\mathbf{x}_j)) \cdot \mathbf{I}_{fg}(\mathbf{x}_{i})
% \end{equation}

\paragraph{Training Parameters:} 
We use the pre-trained FlowNetSimple \citep{dosovitskiy2015flownet} to initialize the weights in our network. Weights of the new layers are randomly initialized, except for the additional weights in the first conv layer that deals with the input masks and the fully-connected layer that predicts the translation, which are initialized with zeros. Other than predicting the pose transformation, the network also predicts the optical flow and the foreground mask. Including the two additional losses could slightly increase the pose estimation performance and make the training more stable. Specifically, we use the optical flow loss $L_\text{flow}$ as in FlowNet \citep{dosovitskiy2015flownet} and the sigmoid cross-entropy loss as the mask loss $L_\text{mask}$. Two deconvolutional blocks in FlowNet are inherited to produce the feature map used for the mask and the optical flow prediction, whose spatial scale is 0.0625. Two $1\times 1$ convolutional layers with output channel 1 (mask prediction) and 2 (flow prediction) are appended after this feature map. The predictions are then bilinearly up-sampled to the original image size ($480\times 640$) to compute losses. 

The overall loss is $L = \alpha L_\text{pose} + \beta L_\text{flow} + \gamma L_\text{mask}$, where we use $\alpha = 0.1$, $\beta = 0.25$, $\gamma = 0.03$ throughout the experiments (except some of our ablation studies). Each training batch contains 16 images. We train the network with 4 GPUs where each GPU processes 4 images. We generate 4 items for each image as described in Sec.~\ref{sec:zoom}: two images and two masks. The observed mask is randomly dilated with no more than 10 pixels to avoid over-fitting.

\paragraph{The Distribution of Rendered Pose during Training:}
%\label{sec:appendix_pose_imgn}
The \rend\ image $\xrend$ and mask $\mrend$ are randomly generated during training without using prior knowledge of the initial poses in the test set. 
Specifically, given a ground truth pose $\gtpose$, we add noises to $\gtpose$ to generate the \rend\ poses. 
For rotation, we independently add a Gaussian noise $\mathcal{N}(0, 15^2)$ to each of the three Euler angles of the rotation. If the angular distance between the new pose and the ground truth pose is more than $45\degree$, we discard the new pose and generate another one in order to make sure the initial pose for refinement is within $45\degree$ of the ground truth pose during training. 
For translation, considering the fact that RGB-based pose estimation methods usually have larger standard deviation on depth estimation, the following Gaussian noises are added to the three components of the translation: $\Delta x \sim \mathcal{N}(0, 0.01^2), \Delta y \sim \mathcal{N}(0, 0.01^2), \Delta z \sim \mathcal{N}(0, 0.05^2)$, where the standard deviations are 1 cm, 1 cm and 5 cm, respectively.

%\label{sec:appendix_data_syn}

\begin{figure*}[t]
\subfloat[Synthetic Data for LINEMOD\label{subig.data_syn_lm}]{
\includegraphics[width=0.30\textwidth]{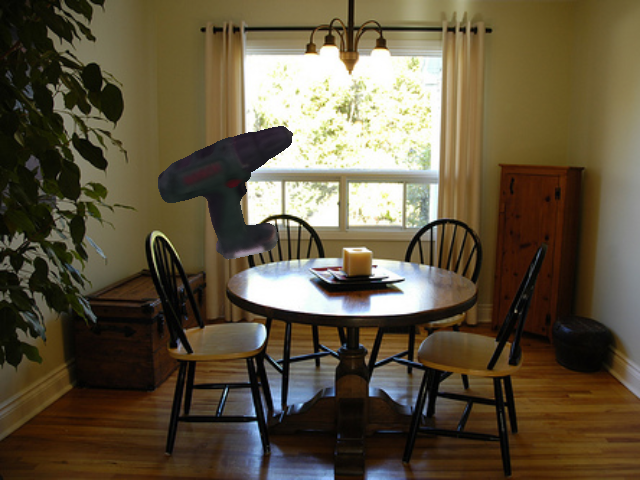}}
\hfill
\subfloat[Synthetic Data for Occlusion LINEMOD\label{subfig.data_syn_occ}]{
\includegraphics[width=0.30\textwidth]{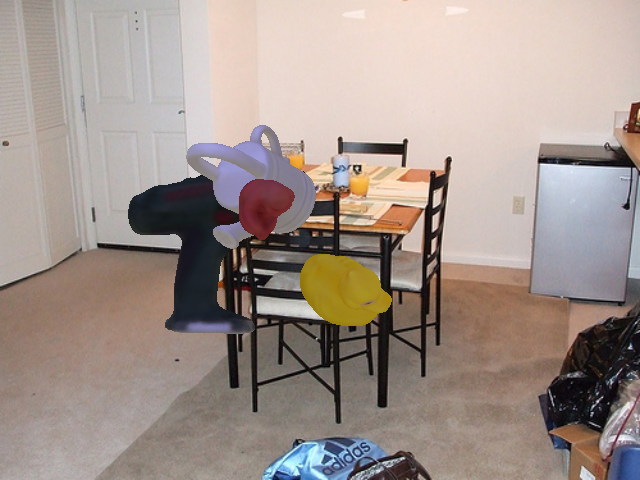}}
\hfill
\subfloat[Synthetic Data for YCB-Video\label{subfig.data_syn_ycb}]{
\includegraphics[width=0.30\textwidth]{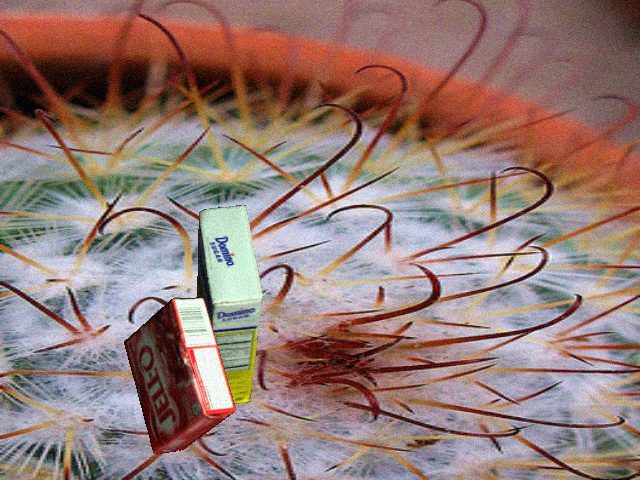}}
\caption{Synthetic Data for the LINEMOD, Occlusion LINEMOD and YCB-Video separately. \ref{subig.data_syn_lm} shows the synthetic training data used when training on the LINEMOD dataset, only one object is presented in the image so there is no occlusion. \ref{subfig.data_syn_occ} shows the synthetic training data used when training on the Occlusion LINEMOD dataset, multiple objects are presented in one image so one object may be occluded by other objects. \ref{subfig.data_syn_ycb} shows the synthetic training data used when training on the YCB-Video dataset. These images are rendered on the fly, so we only render two objects to maintain efficiency.}
\label{fig.data_syn}
\vspace{-4mm}
\end{figure*}

\paragraph{Synthetic Training Data:}
Real training images provided in existing datasets may be highly correlated or lack images in certain situations such as occlusions between objects. Therefore, generating synthetic training data is essential to enable the network to deal with different scenarios in testing. In generating synthetic training data for the LINEMOD dataset, considering the fact that the elevation variation is limited in this dataset, we calculate the elevation range of the objects in the provided training data. Then we rotate the object model with a randomly generated quaternion and repeat it until the elevation is within this range. The translation is randomly generated using the mean and the standard deviation computed from the training set. During training, the background of the synthetic image is replaced by a randomly chosen indoor image from the PASCAL VOC dataset as shown in Fig.~\ref{fig.data_syn}.

For the Occlusion LINEMOD dataset, multiple objects are rendered into one image in order to introduce occlusions among objects. 
The number of objects ranges from 3 to 8 in these synthetic images. 
As in the LINEMOD dataset, the quaternion of each object is also randomly generated to ensure that the elevation range is within that of training data in the Occlusion LINEMOD dataset. 
The translations of the objects in the same image are drawn according to the distributions of the objects in the YCB-Video dataset~\citep{xiang2017posecnn} by adding a small Gaussian noise.

For the YCB-Video dataset, synthetic images are generated on the fly. Other than the target object, we also render another object close to it to introduce partial occlusion.

The real training images may also lack variations in light conditions exhibited in the real world or in the testing set. 
Therefore, we add a random light condition to each synthetic image in both the LINEMOD dataset and the Occlusion LINEMOD dataset.

\subsection{Testing Implementation Details} 
\paragraph{Testing Parameters:}
The mask prediction branch and the optical flow branch are removed during testing. Since there is no ground truth segmentation of the object in testing, we use the tightest bounding box of the rendered mask $\mrend$ instead, so the network searches the neighborhood near the estimated pose to find the target object to match. Unless specified, we use the pose estimates from PoseCNN \citep{xiang2017posecnn} as the initial poses. Our \dimnet\ network runs at 12 fps per object using an NVIDIA 1080 Ti GPU with 2 iterations during testing.

\paragraph{Pose Initialization during inference:}
Our framework takes an input image and an initial pose estimation of an object in the image as inputs, and then refine the initial pose iteratively. In our experiments, we have tested two pose initialization methods.

The first one is PoseCNN~\citep{xiang2017posecnn}, a convolutional neural network designed for 6D object pose estimation. PoseCNN performs three tasks for 6D pose estimation, i.e., semantic labeling to classify image pixels into object classes, localizing the center of the object on the image to estimate the 3D translation of the object, and 3D rotation regression. In our experiments, we use the 6D poses from PoseCNN as initial poses for pose refinement.

To demonstrate the robustness of our framework on pose initialization, we have implemented a simple 6D pose estimation method for pose initialization, where we extend the Faster R-CNN framework designed for 2D object detection~\citep{faster} to 6D pose estimation. Specifically, we use the bounding box of the object from Faster R-CNN to estimate the 3D translation of the object. The center of the bounding box is treated as the center of the object. The distance of the object is estimated by maximizing the overlap of the projection of the 3D object model with the bounding box. To estimate the 3D rotation of the object, we add a rotation regression branch to Faster R-CNN as in PoseCNN. In this way, we can obtain a 6D pose estimation for each detected object from Faster R-CNN.

In our experiments on the LINEMOD dataset described in Sec.~\ref{sec.exp_on_LM}, we have shown that, although the initial poses from Faster R-CNN are much worse than the poses from PoseCNN, our framework is still able to refine these poses using the \emph{same} weights. The performance gap between using the two different pose initialization methods is quite small, which demonstrates the ability of our framework in using different methods for pose initialization. 

\subsection{Evaluation Metrics}
\label{sec:appendix_eval_metrics}
We use the following three evaluation metrics for 6D object pose estimation. i) The \emph{5\degree, 5cm} metric considers an estimated pose to be correct if its rotation error is within 5\degree \ and the translation error is below 5cm. ii) The \emph{6D Pose} metric~\citep{hinterstoisser2012accv} computes the average distance between the 3D model points transformed using the estimated pose and the ground truth pose. For symmetric objects, we use the closest point distance in computing the average distance. An estimated pose is correct if the average distance is within 10\% of the 3D model diameter. iii) The \emph{2D Projection} metric computes the average distance of the 3D model points projected onto the image using the estimated pose and the ground truth pose. An estimated pose is correct if the average distance is smaller than 5 pixels. 
\paragraph{k\degree, k cm:} Proposed in~\cite{Shotton2013cvpr}. The 5\degree, 5cm metric considers an estimated pose to be correct if its rotation error is within 5\degree \ and the translation error is below 5cm. We also provided the results with 2\degree, 2cm and 10\degree, 10cm in Table \ref{table.different_thresholds} to give a comprehensive view about the performance.

For symmetric objects such as eggbox and glue in the LINEMOD dataset, we compute the rotation error and the translation error against all possible ground truth poses with respect to symmetry and accept the result when it matches one of these ground truth poses.

\paragraph{6D Pose:} \cite{hinterstoisser2012accv} use the average distance (ADD) metric to compute the averaged distance between points transformed using the estimated pose and the ground truth pose as in Eq. \ref{eq.add}:
\begin{equation}
\label{eq.add}
\textbf{ADD}=\frac{1}{m}\sum_{x\in \mathcal{M}}\|(\mathbf{R}\mathbf{x}+\mathbf{t}) - (\mathbf{\hat{R}}\mathbf{x}+\mathbf{\hat{t}})\|_2,
\end{equation}
where $m$ is the number of points on the 3D object model, $\mathcal{M}$ is the set of all 3D points of this model, $\pose=[\mathbf{R}|\mathbf{t}]$ is the ground truth pose and $\gtpose=[\mathbf{\hat{R}}|\mathbf{\hat{t}}]$ is the estimated pose. \yi{Here the number of points $m$ can be different from the number of points $n$ used in Eq. \ref{eq.reproj3d} as the point clouds used for training is a subset randomly sampled from the original point clouds to reduce the time to compute the loss during training.} $\mathbf{R}\mathbf{x}+\mathbf{t}$ indicates transforming the point with the given SE(3) transformation (pose) $\pose$. Following~\citep{brachmann2016uncertainty}, we compute the distance between all pairs of points from the model and regard the maximum distance as the diameter $d$ of this model. Then a pose estimation is considered to be correct if the computed average distance is within 10\% of the model diameter. In addition to using $0.1 d$ as the threshold, we also provided pose estimation accuracy using thresholds $0.02 d$ and $0.05 d$ in Table \ref{table.different_thresholds}. We use $0.1 d$ as the threshold of 6D Pose metric in the following paper if not specified.

For symmetric objects, we use the closest point distance in computing the average distance for 6D pose evaluation as in~\cite{hinterstoisser2012accv}:
\begin{equation}
\label{eq.adds}
\textbf{ADD-S}=\frac{1}{m}\sum_{\mathbf{x}_1\in \mathcal{M}} \min_{\mathbf{x}_2 \in \mathcal{M}} \|(\mathbf{R}\mathbf{x}_1+\mathbf{t}) - (\mathbf{\hat{R}}\mathbf{x}_2+\mathbf{\hat{t}})\|_2.
\end{equation}

In the YCB-Video Dataset, we use the metric ADD and ADD-S described in \cite{xiang2017posecnn}. After getting the ADD and ADD-S distance described in Eq.~\ref{eq.add} and Eq.~\ref{eq.adds}, we vary the threshold from 0 to 10 cm and accumulate the area under the accuracy curves.

\paragraph{2D Projection:} focuses on the matching of pose estimation on 2D images. This metric is considered to be important for applications such as augmented reality. We compute the error using Eq. \ref{eq.proj2d} and accept a pose estimation when the 2D projection error is smaller than a predefined threshold:
\begin{equation}
\label{eq.proj2d}
\textbf{Proj. 2D}=\frac{1}{m}\sum_{x\in \mathcal{M}}\|\mathbf{K}(\mathbf{R}\mathbf{x}+\mathbf{t}) - \mathbf{K}(\mathbf{\hat{R}}\mathbf{x}+\mathbf{\hat{t}})\|_2,
\end{equation}
where $\mathbf{K}$ denotes the intrinsic parameter matrix of the camera and $\mathbf{K}(\mathbf{R}\mathbf{x}+\mathbf{t})$ indicates transforming a 3D point according to the SE(3) transformation and then projecting the transformed 3D point onto the image. In addition to using 5 pixels as the threshold, we also show our results with the thresholds 2 pixels and 10 pixels. We use 5 pixels as the threshold of Proj. 2D metric in the following paper if not specified.

For symmetric objects such as eggbox and glue in the LINEMOD dataset, we compute the 2D projection error against all possible ground truth poses and accept the result when it matches one of these ground truth poses.

\subsection{Experiments on the LINEMOD Dataset}
\label{sec.exp_on_LM}
The LINEMOD dataset contains 15 objects. We train and test our method on 13 of them as other methods in the literature. We follow the procedure in \citep{brachmann2016uncertainty} to split the dataset into the training and test sets, with around 200 images for each object in the training set and 1,000 images in the test set. \yi{Fig.~\ref{fig.occ_demo} shows a subset of objects used in LINEMOD dataset. These objects are textureless and thus difficult for pose estimation methods using only local features.}

\paragraph{Training strategy:}%
For every image, we generate 10 random poses near the ground truth pose, resulting in 2,000 training samples for each object in the training set. Furthermore, we generate 10,000 synthetic images for each object where the pose distribution is similar to the real training set. For each synthetic image, we generate 1 random pose near its ground truth pose. Thus, we have a total of 12,000 training samples for each object in training. The background of a synthetic image is replaced with a randomly chosen indoor image from PASCAL VOC~\citep{everingham2010pascal}. We train the networks for 8 epochs with initial learning rate 0.0001. The learning rate is divided by 10 after the 4th and 6th epoch, respectively.

\paragraph{Ablation study on iterative training and testing:}%
Table \ref{table.ablation_iter_size} shows the results that use different numbers of iterations during training and testing. The networks with $train\_iter=1$ and $train\_iter=2$ are trained with 32 and 16 epochs respectively to keep the total number of updates the same as $train\_iter=4$. The table shows that without iterative training ($train\_iter=1$), multiple iteration testing does not improve, potentially even making the results worse ($test\_iter=4$). We believe that the reason is due to the fact that the network is not trained with enough \rend\ poses close to their ground truth poses. The table also shows that one more iteration during training and testing already improves the results by a large margin. The network trained with 2 iterations and tested with 2 iterations is slightly better than the one trained with 4 iterations and tested with 4 iterations. This may be because the LINEMOD dataset is not sufficiently difficult to generate further improvements by using 3 or 4 iterations. Since it is not straightforward to determine how many iterations to use in each dataset, we use 4 iterations during training and testing in all other experiments.

\setlength{\tabcolsep}{4.0pt}
\begin{table*}[t]
\begin{center}
\caption{Ablation study of the number of iterations during training and testing.}
\label{table.ablation_iter_size}

\begin{tabular}{l|c|c|c|c|c|c|c|c|c|c}
\hline
train iter & \multirow{2}{*}{init} & \multicolumn{3}{c|}{1} & \multicolumn{3}{c|}{2} & \multicolumn{3}{c}{4} \\
\cline{1-1} 
\cline{3-11}
test iter & & 1 & 2 & 4 & 1 & 2 & 4 & 1 & 2 & 4\\
\hline
5cm 5\degree& 19.4 & 57.4 & 58.8 & 54.6 & 76.3 & 86.2 & 86.7 & 70.2 & 83.7 & 85.2 \\
6D Pose	& 62.7 & 77.9 & 79.0 & 76.1 & 83.1 & 88.7 & 89.1 & 80.9 & 87.6 & 88.6 \\
Proj. 2D	& 70.2 & 92.4 & 92.6 & 89.7 & 96.1 & 97.8 & 97.6 & 94.6 & 97.4 & 97.5 \\
\hline
\end{tabular}
\end{center}
\end{table*}
\setlength{\tabcolsep}{1.4pt}
% \setlength{\tabcolsep}{4.0pt}
% \renewcommand{\arraystretch}{1.2}
% \begin{table}[t]
% \centering
% \caption{Comparisons with state-of-the-art 6D Pose Estimation methods}
% 
% \begin{tabular}{l|c|c|c|c|c|c|c|c|c|c|c|c|c}
% \scriptsize{methods} & \scriptsize{ape} & \scriptsize{bench.} & \scriptsize{can} & \scriptsize{cat} & \scriptsize{driller} & \scriptsize{duck} & \scriptsize{duck} & \scriptsize{eggbox} & \scriptsize{glue} & \scriptsize{hole.} & \scriptsize{iron} & \scriptsize{lamp} & \scriptsize{phone}\\
% \hline
% \end{tabular}
% \label{table.ablation_iter_size}
% \end{table}

\paragraph{Ablation study on the zoom in strategy, network structures, transformation representations, and loss functions:}%
Table~\ref{table.ablation_to_design} summarizes the ablation studies on various aspects of \dimnet. The ``zoom'' column indicates whether the network uses full images as its input or zoomed in bounding boxes up-sampled to the original image size. Comparing rows 5 and 7 shows that the higher resolution achieved via zooming in provides very significant improvements. 

``Regressor'': We train the \dimnet\ network jointly over all objects, generating a pose transformation independent of the specific input object (labeled ``shared'' in ``regressor'' column). Alternatively, we could train a different 6D pose regressor for each individual object by using a separate fully connected layer for each object after the final FC256 layer shown in Fig.~\ref{fig:network}.  This setting is labeled as ``sep.'' in Table~\ref{table.ablation_to_design}. Comparing rows 3 and 7 shows that both approaches provide nearly indistinguishable results. But the shared network provides some efficiency gains.

``Network'': Similarly, instead of training a single network over all objects, we could train separate networks, one for each object as in~\cite{rad2017bb8}. Comparing row 1 to 7 shows that a single, shared network provides better results than individual ones, which indicates that training on multiple objects can help the network learn a more general representation for matching. \yi{We also present an ablation study of mask prediction and flow prediction in Table \ref{table.ablation_to_mask_and_flow}. It shows that when trained with these two auxiliary branches, the network could achieve the highest performance.}

\begin{table*}[t]
\centering
\caption{Ablation study on the role of mask prediction and flow prediction branch. The networks are trained 5 times for each setting on the object ape of the LINEMOD dataset. The numbers denote mean $\pm$ standard deviation.}
\begin{tabular}{c|c|c|c|c}
\hline
\multicolumn{2}{c|}{methods} & \multirow{2}{*}{5cm 5\degree} & \multirow{2}{*}{6D Pose} & \multirow{2}{*}{Proj.~2D} \\
\cline{1-2}
mask & flow &&& \\
\hline
\checkmark & \checkmark & 93.9$\pm$0.7 & 82.5$\pm$1.7 & 98.2$\pm$0.3 \\
\checkmark & & 91.7$\pm$0.4 & 82.5$\pm$1.6 & 97.7$\pm$0.1 \\
& \checkmark & 89.2$\pm$2.1 & 63.7$\pm$3.4 & 98.4$\pm$0.2 \\
& & 89.6$\pm$0.8 & 72.3$\pm$1.1 & 98.1$\pm$0.1 \\
\hline
\end{tabular}
\label{table.ablation_to_mask_and_flow}
\end{table*}

``Coordinate'': This column investigates the impact of our choice of coordinate frame to reason about object transformations, as described in Fig.~\ref{fig.rot_reprentations}. 
The row labeled ``camera'' provides results when choosing the camera frame of reference as the representation for the object pose, 
rows labeled ``model'' move the center of rotation to the object model and choose the object model coordinate frame to reason about rotations, 
and the ``disentangled'' rows provide our disentangled approach of moving the center into the object model while keeping the camera coordinate frame for rotations. Comparing rows 2 and 3 shows that reasoning in the camera rotation frame provides slight improvements. 
Furthermore, it should be noted that only our ``disentangled'' approach is able to operate on unseen objects.  
Comparing rows 4 and 5 shows the large improvements our representation achieves over the common approach of reasoning fully in the camera frame of reference. 

``Loss'': The traditional loss for pose estimation is specified by the distance (``Dist'') between the estimated and ground truth 6D pose coordinates, i.e., angular distance for rotation and euclidean distance for translation. Comparing rows 6 and 7 indicates that our point matching loss (``PM'') provides significantly better results especially on the 6D pose metric, which is the most important measure for reasoning in 3D space. 

\setlength{\tabcolsep}{4.0pt}
\renewcommand{\arraystretch}{1.2}
\begin{table*}[t]
\centering
\caption{Ablation study on different design choices of the \dimnet\ network on the LINEMOD dataset.}

\begin{tabular}{c|c|c|c|c|c|c|c|c}
\hline
\multirow{2}{*}{Row} & \multicolumn{5}{c|}{methods} & \multirow{2}{*}{5cm 5\degree} & \multirow{2}{*}{6D Pose} & \multirow{2}{*}{Proj.~2D}\\
\cline{2-6}
& zoom & \tabincell{c}{regressor} & \tabincell{c}{network} & \tabincell{c}{coordinate} &\tabincell{c}{loss}& & & \\
\hline
 1 & \checkmark & - 	 & sep.   & disentangled 	& PM	& 83.3 & 87.6 & 96.2 \\
 2 & \checkmark & sep. 	 & shared & model 	    & PM	& 79.2 & 87.5 & 95.4 \\
 3 & \checkmark & sep. 	 & shared & disentangled 	& PM	& 86.6 & 89.5 & 96.7 \\
 4 &            & shared & shared & camera	    & PM	& 16.6 & 44.3 & 62.5 \\
 5 &            & shared & shared & disentangled	& PM    & 38.3 & 65.2 & 80.8 \\
 6 & \checkmark & shared & shared & disentangled 	& Dist  & 86.5 & 79.2 & 96.2 \\
 7 & \checkmark & shared & shared & disentangled 	& PM	& 85.2 & 88.6 & 97.5 \\

\hline
\end{tabular}
\label{table.ablation_to_design}
\end{table*}

\paragraph{Application to different initial pose estimation networks:}

Table~\ref{table.robust_to_initial_pose} provides results when we initialize \dimnet\ with two different pose estimation networks. The first one is PoseCNN~\citep{xiang2017posecnn}, and the second one is  a simple 6D pose estimation method based on Faster R-CNN~\citep{faster}. Specifically, we use the bounding box of the object from Faster R-CNN to estimate the 3D translation of the object. The center of the bounding box is treated as the center of the object. The distance of the object is estimated by maximizing the overlap of the projection of the 3D object model with the bounding box. To estimate the 3D rotation of the object, we add a rotation regression branch to Faster R-CNN as in PoseCNN. As we can see in Table~\ref{table.robust_to_initial_pose}, our network achieves very similar pose estimation accuracy even when initialized with the estimates from the extension of Faster R-CNN, which are not as accurate as those provided by PoseCNN~\citep{xiang2017posecnn}.

%Notice that the same network is used for refinement here with the two types of initial poses.

\setlength{\tabcolsep}{4.0pt}
\renewcommand{\arraystretch}{1.2}
\begin{table}[t]
\centering
\caption{Ablation study on two different methods for generating initial poses on the LINEMOD dataset.}

\begin{tabular}{l|c|c|c|c}
\hline
method 		& PoseCNN & \tabincell{c}{PoseCNN\\+OURS} & \tabincell{c}{Faster\\R-CNN}  & \tabincell{c}{Faster R-CNN\\+OURS}\\
\hline
5cm 5\degree& 19.4 & 85.2 & 11.9 & 83.4\\
6D Pose  	& 62.7 & 88.6 & 33.1 & 86.9\\
Proj. 2D	& 70.2 & 97.5 & 20.9 & 95.7 \\
\hline
\end{tabular}
\label{table.robust_to_initial_pose}
\end{table}

\paragraph{Comparison with the state-of-the-art 6D pose estimation methods:}%
Table~\ref{table.LM_comparison_with_others} shows the comparison with the best color-only techniques on the LINEMOD dataset. \dimnet\ achieves very significant improvements over all prior methods, even those that also deploy refinement steps (BB8~\citep{rad2017bb8} and SSD-6D~\citep{kehl2017ssd}). 

\setlength{\tabcolsep}{4.0pt}
\renewcommand{\arraystretch}{1.2}
\begin{table*}[t]
\centering
\caption{Comparison with state-of-the-art methods on the LINEMOD dataset}
% 
% \footnotesize
\begin{tabular}{l|c|c|c|c|c|c}
\hline
methods & \tabincell{c}{\citeauthor{brachmann2016uncertainty}\\(2016)} & \tabincell{c}{BB8 w/ ref.\\(\citeauthor{rad2017bb8}\\\citeyear{rad2017bb8})} & \tabincell{c}{SSD-6D w\/ ref.\\\citep{kehl2017ssd}} & \tabincell{c}{\citeauthor{tekin2017real}\\(\citeyear{tekin2017real})} & \tabincell{c}{PoseCNN \\~\citep{xiang2017posecnn}} & \tabincell{c}{PoseCNN\\\citep{xiang2017posecnn}\\+OURS}\\
\hline
5cm 5\degree& 40.6 & 69.0 & - 	& - 	& 19.4 & \textbf{85.2} \\
6D Pose		& 50.2 & 62.7 & 79  & 55.95 & 62.7 & \textbf{88.6} \\
Proj. 2D	& 73.7 & 89.3 & -   & 90.37	& 70.2 & \textbf{97.5} \\
\hline
\end{tabular}
\label{table.LM_comparison_with_others}
\end{table*}

\paragraph{Detailed Results on the LINEMOD Dataset:}

Table \ref{table.different_thresholds} shows our detailed results on all the 13 objects in the LINEMOD dataset. The network is trained and tested with 4 iterations and 8 epochs. 
Initial poses are estimated by PoseCNN \citep{xiang2017posecnn}.

\begin{table*}
\centering
\caption{Results of using more detailed thresholds on the LINEMOD dataset}
\small
\begin{tabular}{l|c|c|c|c|c|c|c|c|c}
\hline
\multirow{2}{*}{\tabincell{c}{metric\\threshold}} & \multicolumn{3}{c|}{(n\degree,  n cm)}& \multicolumn{3}{c|}{6D Pose}& \multicolumn{3}{c}{Projection 2D}\\
\cline{2-10}
		 	& (2, 2)&(5, 5) &(10,10)& 0.02d & 0.05d & 0.10d & 2 px. & 5 px. & 10 px.\\
\hline
ape			& 37.7 & 90.4 & 98.0 & 14.3 & 48.6 & 77.0 & 92.2 & 98.4 & 99.6 \\
benchvise	& 37.6 & 88.7 & 98.2 & 37.5 & 80.5 & 97.5 & 67.7 & 97.0 & 99.6 \\
camera		& 56.1 & 95.8 & 99.2 & 30.9 & 74.0 & 93.5 & 86.3 & 98.9 & 99.7 \\
can			& 58.0 & 92.8 & 99.0 & 41.4 & 84.3 & 96.5 & 98.6 & 99.7 & 99.8 \\
cat			& 33.5 & 87.6 & 97.8 & 17.6 & 50.4 & 82.1 & 88.4 & 98.7 & 100.0 \\
driller		& 49.4 & 92.9 & 99.1 & 35.7 & 79.2 & 95.0 & 64.2 & 96.1 & 99.4 \\
duck		& 30.8 & 85.2 & 98.5 & 10.5 & 48.3 & 77.7 & 88.1 & 98.5 & 99.8 \\
eggbox		& 32.1 & 63.9 & 94.5 & 34.7 & 77.8 & 97.1 & 53.4 & 96.2 & 99.6 \\
glue		& 32.8 & 83.0 & 98.0 & 57.3 & 95.4 & 99.4 & 81.5 & 98.9 & 99.7 \\
holepuncher	& 8.7  & 54.5 & 93.8 & 5.3 	& 27.3 & 52.8 & 59.1 & 96.3 & 99.5 \\
iron		& 47.5 & 92.7 & 99.3 & 47.9 & 86.3 & 98.3 & 67.4 & 97.2 & 99.9 \\
lamp		& 47.5 & 90.9 & 98.4 & 45.3 & 86.8 & 97.5 & 60.0 & 94.2 & 99.0 \\
phone		& 34.8 & 89.6 & 98.6 & 22.7 & 60.5 & 87.7 & 75.9 & 97.7 & 99.8 \\
\hline
MEAN	& 39.0 & 85.2 & 97.9 & 30.9 & 69.2 & 88.6 & 75.6 & 97.5 & 99.7 \\
\hline
\end{tabular}
\label{table.different_thresholds}
\end{table*}

% \textbf{Evaluation using different thresholds}\\
% Table. \ref{table.different_thresholds}

\subsection{Experiments on the Occlusion LINEMOD Dataset}
\label{sec.exps_on_occ}
The Occlusion LINEMOD dataset proposed in~\cite{Brachmann2014Learning6O} shares the same images used in the LINEMOD dataset~\citep{hinterstoisser2012accv}, but annotated 8 objects in one video that are heavily blocked by other objects.

\paragraph{Training:} For every real image, we generate 10 random poses as described in Sec.~\ref{sec.exp_on_LM}. Considering the fact that most of the training data lacks occlusions, we generated about 20,000 synthetic images with multiple objects in each image. By doing so, every object has around 12,000 images which are partially occluded, and a total of 22,000 images for each object in training. We perform the same background replacement and training procedure as in the LINEMOD dataset. 

%The background of each synthetic images is replaced with a randomly chosen indoor image from PASCAL VOC~\citep{everingham2010pascal}.  We train the network with 8 epochs with initial learning rate of 1e-4. The learning rate is divided by 10 after the 4th and 6th epoch, respectively.

\paragraph{Comparison with the state-of-the-art methods:} The comparison between our method and other RGB-only methods is shown in Fig.~\ref{fig.occ_state_of_the_art}. We only show the plots with accuracies on the 2D Projection metric because these are the only results reported in~\cite{rad2017bb8} and~\citep{tekin2017real} (results for eggbox and glue use a symmetric version of this accuracy). It can be seen that our method greatly improves the pose accuracy generated by PoseCNN and surpasses all other RGB-only methods by a large margin. It should be noted that BB8~\citep{rad2017bb8}  achieves the reported results only when using ground truth bounding boxes during testing. Our method is even competitive with the results that use depth information and ICP to refine the estimates of PoseCNN. Fig.~\ref{fig.occ_demo} shows some pose refinement results from our method on the Occlusion LINEMOD dataset.

\paragraph{Detailed Results on the Occlusion LINEMOD Dataset:}
%\label{sec:appendix_occlusion_results}
\begin{table}
\centering
\caption{Results on the Occlusion LINEMOD dataset. The network is trained and tested with 4 iterations.}
\small
\begin{tabular}{l|c|c|c|c|c|c}
\hline
metric & \multicolumn{2}{c|}{(5\degree,  5cm)}& \multicolumn{2}{c|}{6D Pose}& \multicolumn{2}{c}{Projection 2D}\\
\hline
method & Init. & Refined & Init. & Refined & Init. & Refined \\
\hline
ape			& 2.3 & \textbf{51.8} & 9.9 & \textbf{59.2} & 34.6 & \textbf{69.0}  \\
can			& 4.1 & \textbf{35.8} & 45.5 & \textbf{63.5} & 15.1 & \textbf{56.1}  \\
cat			& 0.3 & \textbf{12.8} & 0.8 & \textbf{26.2} & 10.4 & \textbf{50.9}  \\
driller		& 2.5 & \textbf{45.2} & 41.6 & \textbf{55.6} & 7.4 & \textbf{52.9}  \\
duck		& 1.8 & \textbf{22.5} & 19.5 & \textbf{52.4} & 31.8 & \textbf{60.5}  \\
eggbox		& 0.0 & \textbf{17.8} & 24.5 & \textbf{63.0} & 1.9 & \textbf{49.2}  \\
glue		& 0.9 & \textbf{42.7} & 46.2 & \textbf{71.7} & 13.8 & \textbf{52.9}  \\
hole.	& 1.7  & \textbf{18.8} & 27.0 & \textbf{52.5} & 23.1 & \textbf{61.2}  \\
\hline
MEAN	& 1.7 & \textbf{30.9} & 26.9 & \textbf{55.5} & 17.2 & \textbf{56.6}  \\
\hline
\end{tabular} \label{table.result_on_occlusion}
\end{table}

\begin{figure*}[t]
	\centering
	\includegraphics[width = .8\linewidth]{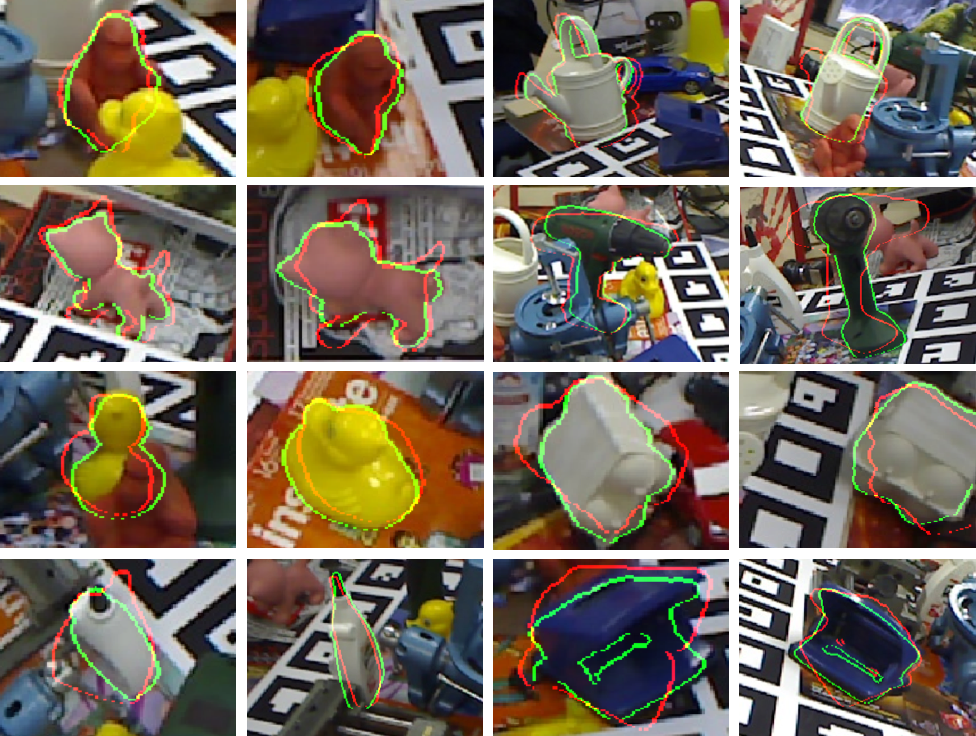}
	\caption{\small{Some pose refinement results on the Occlusion LINEMOD dataset. The red and green lines represent the edges of 3D model projected from the initial poses and our refined poses respectively.}}
	\label{fig:occ_res}
\end{figure*}

Table \ref{table.result_on_occlusion} shows our results on the Occlusion LINEMOD dataset. We can see that DeepIM can significantly improve the initial poses from PoseCNN. 
Notice that the diameter here is computed using the extents of the 3D model following the setting of~\citep{xiang2017posecnn} and other RGB-D based methods. Some qualitative results are shown in Figure \ref{fig:occ_res}.

\if 0
\begin{figure*}[!ht]
\subfloat[ape \label{subig.occ_ape}]{
\includegraphics[width=0.24\textwidth]{7_ape_arp2d_iter4.png}}
\hfill
\subfloat[can\label{subig.occ_can}]{
\includegraphics[width=0.24\textwidth]{7_can_arp2d_iter4.png}}
\hfill
\subfloat[cat\label{subig.occ_cat}]{
\includegraphics[width=0.24\textwidth]{7_cat_arp2d_iter4.png}}
\hfill
\subfloat[driller\label{subig.occ_driller}]{
\includegraphics[width=0.24\textwidth]{7_driller_arp2d_iter4.png}}
\hfill
\subfloat[duck\label{subig.occ_duck}]{
\includegraphics[width=0.24\textwidth]{7_duck_arp2d_iter4.png}}
\hfill
\subfloat[eggbox\label{subig.occ_eggbox}]{
\includegraphics[width=0.24\textwidth]{7_eggbox_arp2d_iter4.png}}
\hfill
\subfloat[glue\label{subig.occ_glue}]{
\includegraphics[width=0.24\textwidth]{7_glue_arp2d_iter4.png}}
\hfill
\subfloat[holepuncher\label{subig.occ_holepuncher}]{
\includegraphics[width=0.24\textwidth]{7_holepuncher_arp2d_iter4.png}}
\hfill
\caption{Comparison with state-of-the-art methods on the Occlusion LINEMOD dataset~\citep{Brachmann2014Learning6O}. Accuracies are measured via the Projection 2D metric.}
\label{fig.occ_state_of_the_art}
\end{figure*}
\fi

\begin{figure*}[t]
	\centering
	\includegraphics[width = \linewidth]{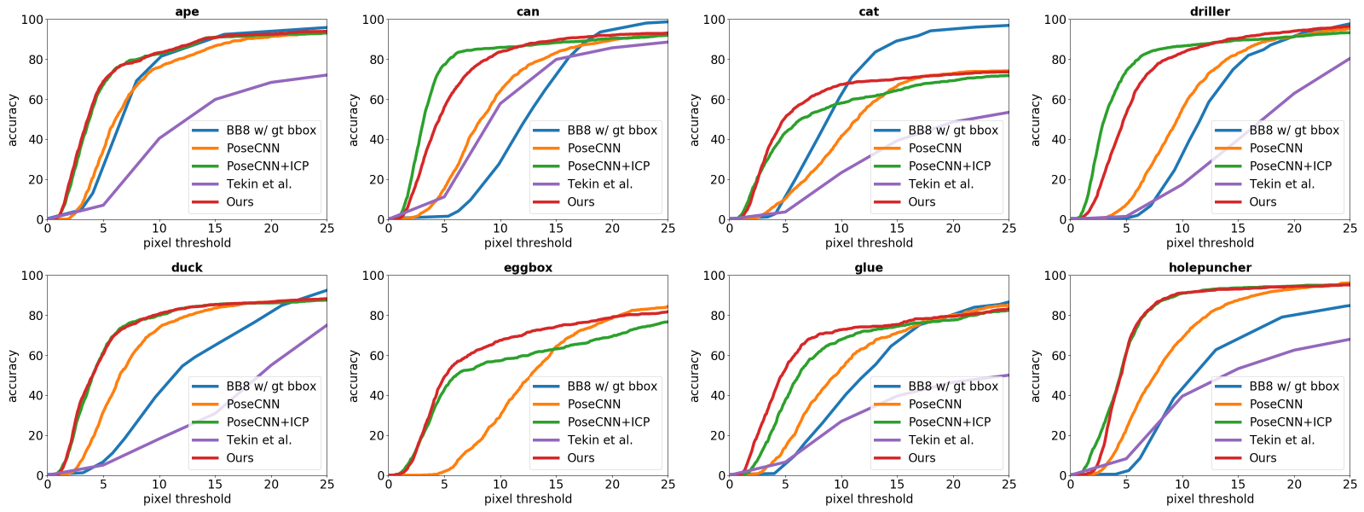}
\caption{Comparison with state-of-the-art methods on the Occlusion LINEMOD dataset~\citep{Brachmann2014Learning6O}. Accuracies are measured via the Projection 2D metric.}
\label{fig.occ_state_of_the_art}
\end{figure*}

\begin{figure*}[t]
	\centering
	\includegraphics[width = .8\linewidth]{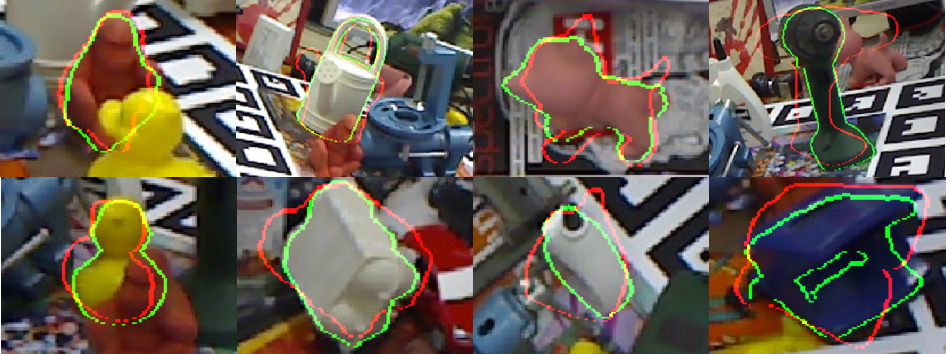}
\caption{Examples of refined poses on the Occlusion LILNEMOD dataset using the results from PoseCNN~\citep{xiang2017posecnn} as initial poses. The red and green lines represent the silhouettes of the initial estimates and our refined poses, respectively.}
\label{fig.occ_demo}
\end{figure*}

\if 0
\begin{figure*}
\subfloat[driller \label{subig.demo_occ_driller}]{
\includegraphics[width=0.28\textwidth]{8_Occ_iter4_results_driller.png}}
\hfill
\subfloat[can\label{subig.demo_occ_can}]{
\includegraphics[width=0.28\textwidth]{8_Occ_iter4_results_can.png}}
\hfill
\subfloat[ape\label{subig.demo_occ_ape}]{
\includegraphics[width=0.28\textwidth]{8_Occ_iter4_results_ape.png}}
\hfill
\caption{Examples of refined poses on the Occlusion LILNEMOD dataset using the results from PoseCNN~\citep{xiang2017posecnn} as initial poses. The red and green lines represent the edges of the initial estimates and our refined poses, respectively.}
\label{fig.occ_demo}
\end{figure*}
\fi

% \setlength{\tabcolsep}{4.0pt}
% \renewcommand{\arraystretch}{1.2}
% \begin{table}[t]
% \centering
% \caption{results of ReProjection 2D metric on the Occlusion dataset. The threshold is 5 pixels}
% 
% \begin{tabular}{l|c|c|c}
% \hline
% methods		& PoseCNN 	& PoseCNN+ICP& PoseCNN+Ours 	\\
% \hline
% ape			& 34.56		& 67.32 & 69.32	\\
% can			& 15.09		& 77.36 & 77.36	\\
% cat			& 10.37		& 43.39 & 43.39	\\
% driller		& 7.36		& 74.36 & 74.36	\\
% duck		& 31.76		& 61.50 & 60.54	\\
% eggbox		& 1.86		& 42.84	& 49.18 \\
% glue		& 13.79		& 37.63	& 52.92 \\
% holepuncher	& 23.06		& 61.32	& 61.16 \\
% \hline
% mean 		& 17.23		& 56.60 & 58.22	\\
% \hline
% \end{tabular}
% \label{table.occLM_comparison_with_others}
% \end{table}
\subsection{Experiments on the YCB-Video Dataset}
\label{sec:ycb_video}
The YCB-Video Dataset, which is proposed in \citep{xiang2017posecnn}, annotates 21 YCB objects~\citep{calli2015ycb} in 92 video sequences (133,827 frames). It is a challenging dataset as the objects have varied sizes
%ATW(diameter from 10 cm to 40 cm)
(diameter from 10 cm to 40 cm), different types of symmetries,
%ATW and the videos contain a large variety of
and a large variety of occlusions and lighting conditions.
We split the dataset as \citep{xiang2017posecnn}, with 80 video sequences for training and 2,949 keyframes in the remaining 12 videos for testing.

\paragraph{Training Strategy:} As images in one video are similar to those in nearby frames, we use 1 image out of every 10 images in the training set for training. Training batches consist of captured real images from the dataset (1/8) and synthetic images which are partially occluded and generated on the fly (7/8). The network is trained with 8 epochs and we decrease the learning rate after 4 and 6 epochs. We found that with large training sets and enough epochs it was not necessary to include the flow prediction and the masks in the input, so we removed those branches and the corresponding loss from this experiment. For different categories, they share the same network but use separate regressors to achieve the best performance.

\paragraph{Evaluation Metric:} We follow the PoseCNN \citep{xiang2017posecnn} paper when evaluating the results which uses accuracy under curve of ADD (Eq. \ref{eq.add}) and ADD-S (Eq. \ref{eq.adds} for each object. We also report the results of ADD(-S) and AUC ADD(-S) metric which is similar to the one we used in LINEMOD~\citep{Brachmann2014Learning6O}. More specifically, we use ADD when the object is not
%ATW symmetry
symmetric
and use ADD-S when the object is
%ATW symmetry
symmetric. Then we compute the averaged accuracy as the final result.

\paragraph{Symmetric Objects:} As described in Sec.~\ref{sec.training_details}, we only keep \rend~ poses that have an angular distance less than 45 degrees from ground truth poses during training, which means we don't need to take special care of objects which have a symmetry angle of more than 90 degrees. However, object 024\_bowl in the YCB-Video dataset is
% ATW rotational
rotational
symmetric. To deal with this kind of symmetry, rather than using the ground truth pose $\gtpose$ provided by the dataset to compute the loss, we choose the distance to the closest pose $\mathbf{p}^*$ among all poses that look the same as the ground truth pose:
\begin{equation}
\centering
    \mathbf{p}^*=\arg\min_{\mathbf{p}\in \mathcal{Q}}\Theta(\mathbf{p}, \mathbf{p}_\text{src})
\end{equation}
Here, $\mathcal{Q}$ denotes the set of poses whose corresponding \rend~ images are the same as the one rendered using the ground truth pose. We assume that the rotation axis goes through the origin of the model frame so that no translation needs to be considered. In the experiment, we calibrate the rotation axis manually and use bisection search to locate the \textit{closest ground truth pose}. Table.~\ref{table.ablation_on_symmetry} compares networks trained with and without this strategy, showing that this training loss is useful.

\begin{table}[t]
    \centering
    \begin{tabular}{c|c|c|c}
    \hline
024\_bowl & init & common & closest \\
\hline
ADD & 54.2 & 55.6 & 68.4 \\
ADD-S & 76.0 & 70.6 & 80.9 \\
\hline
\end{tabular}
    \caption{Ablation study about using \textit{closest ground truth pose} to handle
    %ATW rotational
    rotational
    symmetric objects. These three columns show the evaluation results of initial poses, poses refined by a DeepIM network that treats 024\_bowl as a regular object, and poses refined by a network trained with \textit{closest ground truth pose}. Initial poses are generated as \rend~pose during training described in Sec.~\ref{sec.training_details}}
    \label{table.ablation_on_symmetry}
\end{table}

\begin{table*}[t]
\centering
\caption{Overall results on YCB video results compared with PoseCNN \citep{xiang2017posecnn} and PoseRBPF \citep{deng2019pose}. The ADD(-S) metric and AUC ADD(-S) metric is introduced in Sec.~\ref{sec:ycb_video}}
\small
\begin{tabular}{l|c|c|c|c|c|c|c}
\hline
\multirow{3}{*}{ Methods} & \multicolumn{4}{c|}{RGB} & \multicolumn{3}{c}{RGB-D} \\
\cline{2-8}
&PoseCNN & PoseRBPF++ & \tabincell{c}{PoseCNN\\+\dimnet} & \tabincell{c}{\dimnet\\+Tracking} & \tabincell{c}{PoseCNN\\+ICP} & PoseRBPF & \tabincell{c}{PoseCNN\\+\dimnet} \\
\hline
ADD(-S)\textless 2cm & 27.55 & - & 71.5 & \textbf{79.0} & 78.9 & - & \textbf{90.3}  \\
AUC of ADD(-S) & 61.31 & 64.4 & 81.9 & \textbf{85.9 }& 86.6 & 88.5 & \textbf{90.4} \\
\hline
\end{tabular}
\label{table.ycb_results_simple}
\end{table*}

\begin{table*}[t]
\centering
\small
\caption{Detailed Results on the YCB-Video dataset compared with PoseCNN \citep{xiang2017posecnn} and DenseFusion \citep{wang2019densefusion}. The network is trained and tested with 4 iterations. The ADD and ADD-S is short for AUC of ADD and AUC of ADD-S.}
\begin{tabular}{l|c|c|c|c|c|c|c|c|c|c|c}
% \caption{Detailed Results Comparison with state-of-the-art methods on the YCB-Video dataset.}
\hline
\multirow{3}{*}{ Methods} & \multicolumn{6}{c|}{RGB} & \multicolumn{5}{c}{RGB-D}\\
\cline{2-12}
 & \multicolumn{2}{|c|}{\tabincell{c}{PoseCNN}} & \multicolumn{2}{c|}{\tabincell{c}{PoseCNN\\+\dimnet}} & \multicolumn{2}{c|}{\tabincell{c}{\dimnet\\ Tracking}} & \multicolumn{2}{c|}{\tabincell{c}{PoseCNN\\+ICP}} & \tabincell{c}{DenseFusion} & \multicolumn{2}{c}{\tabincell{c}{PoseCNN\\+\dimnet}}\\
\hline
Evaluation Metric & ADD & ADD-S & ADD & ADD-S & ADD & ADD-S & ADD & ADD-S & ADD-S & ADD & ADD-S\\
\hline
002\_master\_chef\_can & 50.2 & 83.9 & 71.2 & 93.1 & \textbf{89.0} &\textbf{ 93.8} & 68.1 & 95.8 & \textbf{96.4} & \textbf{78.0} & 96.3 \\
003\_cracker\_box & 53.1 & 76.9 & 83.6 & 91.0 &\textbf{ 88.5 }& \textbf{93.0} & 83.4 & 92.7 &\textbf{ 95.5 }& \textbf{91.4} & 95.3 \\
004\_sugar\_box & 68.4 & 84.3 & 94.1 & 96.2 & \textbf{94.3} & \textbf{96.3} & 97.2 & \textbf{98.2} & 97.5 & 97.6 & \textbf{98.2} \\
005\_tomato\_soup\_can & 66.2 & 81.0 & 86.1 & 92.4 & \textbf{89.1} & \textbf{93.2} & 81.8 & 94.5 & 94.6 & \textbf{90.3} & \textbf{94.8} \\
006\_mustard\_bottle & 81.0 & 90.4 & 91.5 & \textbf{95.1} & \textbf{92.0 }& \textbf{95.1} & \textbf{98.0} & 98.6 & 97.2 & 97.1 & \textbf{98.0} \\
007\_tuna\_fish\_can & 70.7 & 88.1 & 87.7 & 96.1 & \textbf{92.0} & \textbf{96.4} & 83.9 & 97.1 & 96.6 & \textbf{92.2} & \textbf{98.0} \\
008\_pudding\_box & 62.7 & 79.1 & \textbf{82.7} & \textbf{90.7} & 80.1 & 88.3 & \textbf{96.6} & \textbf{97.9} & 96.5 & 83.5 & 90.6 \\
009\_gelatin\_box & 75.2 & 87.2 & 91.9 & 94.3 & \textbf{92.0} & \textbf{94.4} & \textbf{98.1} & \textbf{98.8} & 98.1 & 98.0 & 98.5 \\
010\_potted\_meat\_can & 59.5 & 78.5 & 76.2 & 86.4 & \textbf{78.0} &\textbf{ 88.9} & \textbf{83.5} & \textbf{92.7} & 91.3 & 82.2 & 90.3 \\
011\_banana & 72.3 & 86.0 & \textbf{81.2 }& \textbf{91.3} & 81.0 & 90.5 & 91.9 & 97.1 & 96.6 & \textbf{94.9} & \textbf{97.6} \\
019\_pitcher\_base & 53.3 & 77.0 & 90.1 & 94.6 & \textbf{90.4} & \textbf{94.7} & 96.9 & 97.8 & 97.1 & \textbf{97.4} & \textbf{97.9} \\
021\_bleach\_cleanser & 50.3 & 71.6 & 81.2 & 90.3 & \textbf{81.7} &\textbf{ 90.5} & \textbf{92.5} & \textbf{96.9} & 95.8 & 91.6 & \textbf{96.9} \\
\textbf{024\_bowl} & 30.0 & 70.0 & 8.6 & 81.4 & \textbf{38.8} & \textbf{90.6} & \textbf{47.6} & 80.8 & \textbf{88.2} & 8.1 & 87.0 \\
025\_mug & 58.5 & 78.2 & 81.4 & 91.3 & \textbf{83.2} & \textbf{92.0} & 81.1 & 95.0 & 97.1 & \textbf{94.2} &\textbf{ 97.6} \\
035\_power\_drill & 55.3 & 72.7 & \textbf{85.5} & \textbf{92.3} & 85.4 & \textbf{92.3} & \textbf{97.7} & \textbf{98.2} & 96.0 & 97.2 & 97.9 \\
\textbf{036\_wood\_block} & 26.6 & 64.3 & \textbf{60.0} & \textbf{81.9} & 44.3 & 75.4 & 70.9 & 87.6 & 89.7 & \textbf{81.1} & \textbf{91.5} \\
037\_scissors & 35.8 & 56.9 & 60.9 & 75.4 & \textbf{70.3} & \textbf{84.5} & 78.4 & 91.7 & 95.2 & \textbf{92.7} & \textbf{96.0} \\
040\_large\_marker & 58.3 & 71.7 & 75.6 & 86.2 & \textbf{80.4} & \textbf{91.2} & 85.3 & 97.2 & 97.5 & \textbf{88.9} & \textbf{98.2} \\
\textbf{051\_large\_clamp} & 24.6 & 50.2 & 48.4 & 74.3 & \textbf{73.9} & \textbf{84.1} & 52.1 & 75.2 & 72.9 & \textbf{54.2} & \textbf{77.9} \\
\textbf{052\_extra\_large\_clamp} & 16.1 & 44.1 & 31.0 & 73.3 & \textbf{49.3} & \textbf{90.3} & 26.5 & 64.4 & 69.8 & \textbf{36.5} & \textbf{77.8} \\
\textbf{061\_foam\_brick} & 72.9 & 88.2 & 35.9 & 81.9 & \textbf{91.6} &\textbf{95.5} & \textbf{90.5} & 97.4 & 92.5 & 48.2 & \textbf{97.6} \\
\hline
MEAN & 53.4 & 74.6 & 71.7 & 88.1 & \textbf{79.3} & \textbf{91.0} & 80.6 & 92.4 & 93.0 & \textbf{80.7} & \textbf{94.0} \\

\hline
\end{tabular}
\label{table.ycb_results}
\end{table*}

\begin{figure*}[t]
	\centering
	\includegraphics[width = \linewidth]{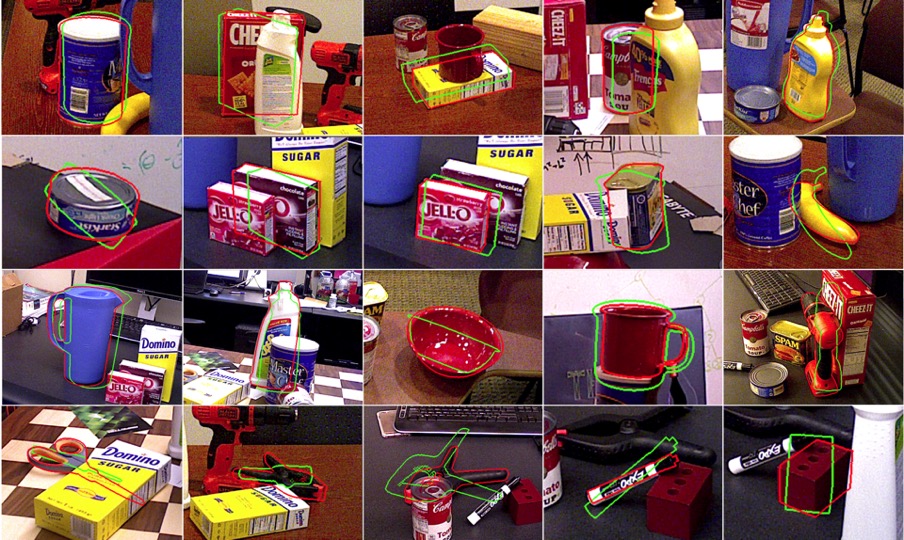}
\caption{Examples of refined poses on the YCB-Video dataset which use results from PoseCNN~\citep{xiang2017posecnn} as initial poses. The \yi{green and red} lines represent the silhouettes of the initial estimates and our refined poses, respectively.}
\label{fig.ycb_demo}
\end{figure*}

\paragraph{Comparison with state-of-the-art methods:}
Table \ref{table.ycb_results} compares our results with two state-of-the-art methods: PoseCNN~\citep{xiang2017posecnn} and DenseFusion~\citep{wang2019densefusion}. As can be seen, \dimnet~ greatly refines the initial pose provided by PoseCNN and is on par with those refined with ICP on many objects despite not using any depth or point cloud data. Notice that DeepIM produces low numbers on symmetric objects, like 024\_bowl, under ADD metric. This is because the ADD metric cannot well represent the performance on symmetric objects as such objects have multiple correct poses but only one of these poses are labeled as the ground truth in the dataset. Table \ref{table.ycb_results_simple} shows the result compared with PoseCNN \citep{xiang2017posecnn} and PoseRBPF \citep{deng2019pose} using the ADD(-S) metrci which can avoid such problems. Fig.~\ref{fig.ycb_demo} visualizes some pose refinement results from our method on the YCB-Video dataset.

\paragraph{Tracking in the YCB-Video Dataset:} Considering the similarity between pose refinement and object tracking, it is natural to use \dimnet~ to track objects in videos. Therefore, we conducted an experiment testing \dimnet 's ability to track objects in the YCB-Video dataset. Provided with the ground truth pose of an object in the first frame of each video, \dimnet~ can perform tracking by using the refined pose estimate from the previous frame as the initial pose of the next frame. Rather than doing inference only on key frames, we applied DeepIM to all images in the test video so that the object poses were close between successive frames. 

In order to determine when DeepIM loses track of an object due to heavy occlusion, we follow a simple strategy: we count the tracking as ``lost'' if the last iteration of the last 10 frames has an average rotation greater than 10 degrees or an average translation greater than 1 cm. Once the tracking is marked as lost, the network will be re-initialzed with PoseCNN's prediction. This strategy is designed with the intuition that  successful
%ATW tracking should have a small offset at the last iteration.
tracking should have a small offset at the last iteration.
Re-initialization happens every 340 frames on average. Table \ref{table.ycb_results_simple} and Table \ref{table.ycb_results} shows our numerical results. Notice that the results of tracking are better than PoseCNN+\dimnet~in most cases and are comparable to the results refined with ICP which uses depth information. Also note that the performance on object 036\_wood\_block is bad because the model of the wooden block is different from the object used in the actual dataset video, which makes it nearly impossible to match the model with the image. 
\begin{figure*}[t]
	\centering
	\includegraphics[width = .8\linewidth]{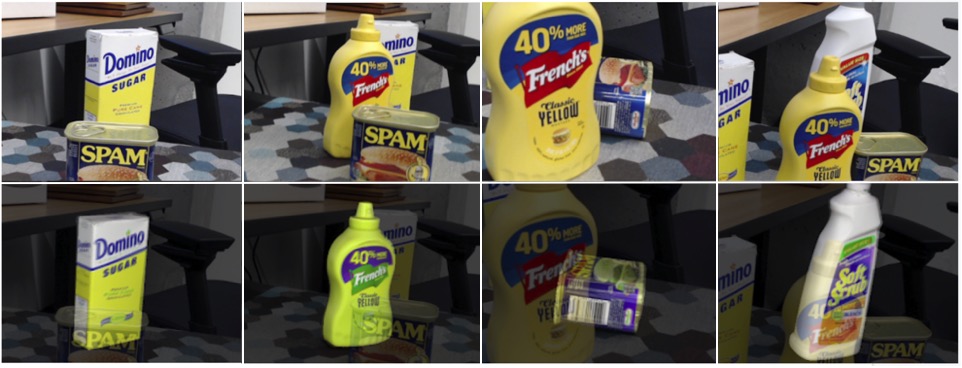}
\caption{Examples on tracking in the real world, using the same network as in Table.~\ref{table.ycb_results} and no prior knowledge about focal length.  The first row shows the images captured with a webcam and the second row renders the object onto the image based on the estimated pose.}
\label{fig.real_tracking}
\end{figure*}
\paragraph{Tracking YCB objects in real scenes:} To demonstrate our framework's generalization, we use our network to track objects in real scenes. This means we don't have any prior knowledge about the
%ATW light condition
lighting conditions, background, or camera parameters. Similar to tracking on the YCB-Video dataset, we use \dimnet~ to refine poses predicted from the
%ATW last
previous frame. Thanks to the disentangled representation, we did not have to calibrate the camera to get its intrinsic matrix. Fig.~\ref{fig.real_tracking} shows some tracking results using our method in the real world environment in real time. 
%Note that compared to methods based on RGB-D images, \dimnet~could directly benefit from higher resolution cameras.
\paragraph{Using Depth information:} Other than using RGB images to do pose refinement, \dimnet can be easily extended to utilize depth information to improve its performance. Here we append the depth images of the observed image and the rendered image with the two zero-initialized additional channels in the first convolution (one for the rendered depth and the other for the observed depth). To provide the network with information of the center of the object, we normalize the depth images by
%ATW subtract them with
subtract them from
the depth of the object's center. The results are shown in Table. \ref{table.ycb_results}.
\begin{figure*}[t]
	\centering
	\includegraphics[width = \linewidth]{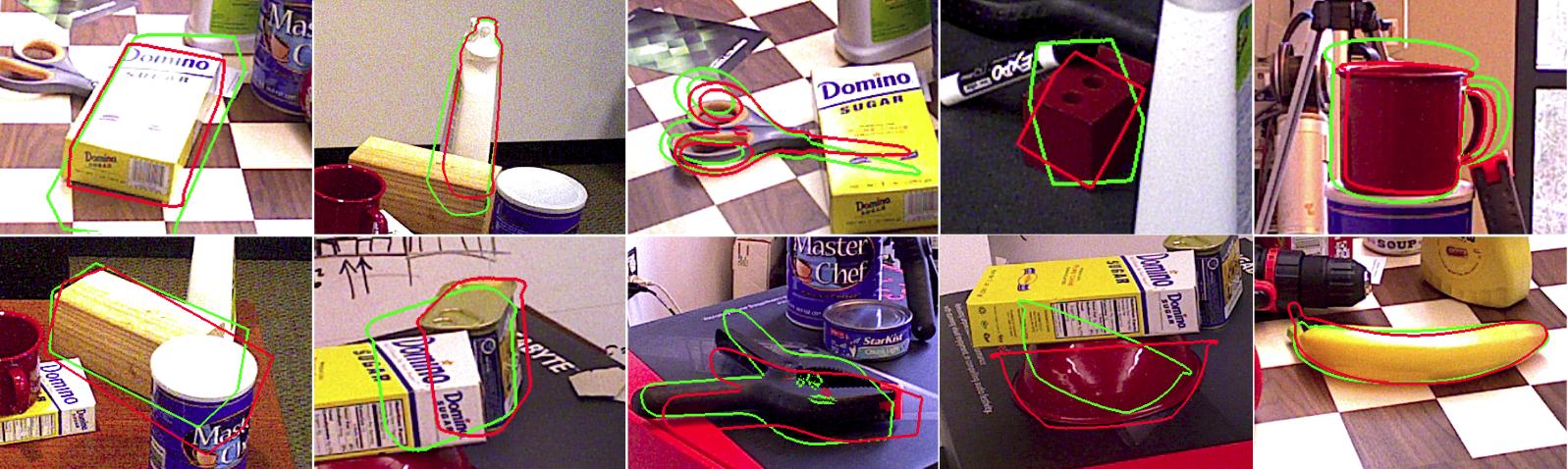}
\caption{Failure cases in YCB-Video dataset. \yi{These images illustrate 5 different reasons we concluded that leads to fail cases.}}
\label{fig.ycb_failure}
\end{figure*}

\paragraph{Failure cases:}
In Fig.~\ref{fig.ycb_failure} we show 10 instances that the network fails to refine to a correct pose. They can be grouped into 5 categories: \yi{1) discrepancy between object models and images.
%ATW It can be caused by bad light conditions
This can be caused by bad light conditions
 or an inaccurate object model; 2) few patterns to match.
%ATW It
This 
usually happens when only certain featureless side-views are visible or the object is heavily
%blocked;
occluded;
3) objects' shapes are unusu    al and difficult to learn; 4) the initial pose is too far away from the correct pose; 5) objects with tiny key components.}

\subsection{Application to Unseen Objects and Unseen Categories}

As stated in Sec.~\ref{sec:untangle}, we designed the disentangled pose representation such that it is independent of the coordinate frame and the size of a specific 3D object model. In other words, the transformation predicted from the network does not need to have prior knowledge about the model itself. Therefore, the pose transformations correspond to operations in the image space. This opens the question whether \dimnet\ can refine the poses of objects that are not included in the training set. From the experiment results we found that our network can perform accurate refinement on these unseen models. See Fig.~\ref{fig.unseen_demo} for example results. We also tested our framework on refining the poses of unseen object categories, where the training categories and the test categories are completely different. 

\begin{figure*}[t]
	\centering
	\includegraphics[width = .8\linewidth]{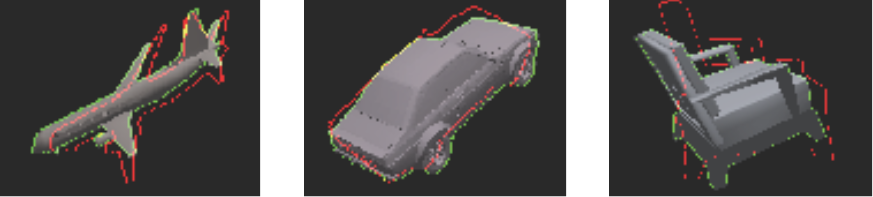}
\caption{Results on pose refinement of 3D models from the ModelNet dataset. These instances were not seen in training. The red and green lines represent the edges of the initial estimates and our refined poses.}
\label{fig.unseen_demo}
\end{figure*}
\paragraph{Test on Unseen Objects:}
%\label{sec:appendix_unseen_objects}

%\setlength{\tabcolsep}{4.0pt}
%\renewcommand{\arraystretch}{1.2}
\begin{table}
\centering
\caption{Results on unseen objects. These models are not included in the training set.}
\small
\begin{tabular}{l|c|c|c|c|c|c}
\hline
category & \multicolumn{2}{c|}{airplane}& \multicolumn{2}{c|}{car}& \multicolumn{2}{c}{chair}\\
\hline
method  & Init. & Refined & Init. & Refined &Init. & Refined\\
\hline
5cm 5\degree	& 0.8 & \textbf{68.9} & 1.0 	& \textbf{81.5} & 1.0 	& \textbf{87.6} \\
6D Pose  & 25.7  & \textbf{94.7} & 10.8	& \textbf{90.7} & 14.6 & \textbf{97.4} \\
Proj. 2D & 0.4 & \textbf{87.3} & 0.2 	& \textbf{83.9} & 1.5	& \textbf{88.6} \\
\hline
\end{tabular}
\label{table.unseen_objects}
\end{table}
% airplane: deepim_v3_flownet_ModelNet_v1_se3_ex_u2s16_airplane_iter_v10_zoom_RFx4_no_pre_mask_v3_camera_real_4epoch_gu
% car: deepim_v3_flownet_ModelNet_v1_se3_ex_u2s16_car_iter_v10_zoom_RFx4_no_pre_mask_v3_camera_real_4epoch_gu_4gpus
% chair: deepim_v3_flownet_ModelNet_v1_se3_ex_u2s16_chair_iter_v10_zoom_RFx4_no_pre_mask_v3_camera_real_4epoch_gu_4gpus

In this experiment, we explore the ability of the network in refining poses of objects that has never been seen in training. 
ModelNet~\citep{wu20153d} contains a large number of 3D models in different object categories. 
Here, we tested our network on three of them: \textit{airplane, car and chair}. For each of these categories, we train a network on no more than 200 3D models and test its performance on 70 unseen 3D models from the same category. Similar to the way that we generate synthetic data as described in Sec \ref{sec.training_details}, we generate 50 poses for each model as the target poses and train the network for 4 epochs. We use uniform gray texture for each model and add a light source which has a fixed relative position to the object to reflect the norms of the object. 
The initial pose used in training and testing is generated in the same way as we did in previous experiments as described in Sec.~\ref{sec.training_details}. The results are show in Table \ref{table.unseen_objects}.

\paragraph{Test on Unseen Categories:}
%\label{sec:appendix_unseen_categories}

%\setlength{\tabcolsep}{4.0pt}
%\renewcommand{\arraystretch}{1.2}
\begin{table}
\centering
\caption{Results on unseen categories. These categories has never been seen by the network during training.}
\small
\begin{tabular}{l|c|c|c|c|c|c}
\hline
metric & \multicolumn{2}{c|}{(5\degree,  5cm)}& \multicolumn{2}{c|}{6D Pose}& \multicolumn{2}{c}{Projection 2D}\\
\hline
method & Init. & Refined & Init. & Refined & Init. & Refined \\
\hline
bathtub			& 0.9 & \textbf{71.6} & 11.9 & \textbf{88.6} & 0.2 & \textbf{73.4}  \\
bookshelf		& 1.2 & \textbf{39.2 }& 9.2 & \textbf{76.4} & 0.1 & \textbf{51.3}  \\
guitar			& 1.2 & \textbf{50.4} & 9.6 & \textbf{69.6} & 0.2 & \textbf{77.1}  \\
range hood		& 1.0 & \textbf{69.8} & 11.2 & \textbf{89.6} & 0.0 & \textbf{70.6}  \\
sofa			& 1.2 & \textbf{82.7} & 9.0 & \textbf{89.5} & 0.1 & \textbf{94.2}  \\
wardrobe		& 1.4 & \textbf{62.7} & 12.5 & \textbf{79.4} & 0.2 & \textbf{70.0}  \\
tv stand		& 1.2 & \textbf{73.6} & 8.8 & \textbf{92.1} & 0.2 & \textbf{76.6}  \\
\hline
\end{tabular} \label{table.unseen_categories}
\end{table}

We also tested our framework on refining the poses of unseen object categories, where the training categories and the test categories are completely different. We train the network on 8 categories from ModelNet~\citep{wu20153d}: \textit{airplane, bed, bench, car, chair, piano, sink, toilet} with 30 models in each category and 50 image pairs for each model. 
The network was trained with 4 iterations and 4 epochs. 
Then we tested the network on 7 other categories: \textit{bathtub, bookshelf, guitar, range hood, sofa, wardrobe, and tv stand}. 
The results are shown in Table. \ref{table.unseen_categories}. 
It shows that the network indeed has learned some general features for pose refinement across different object categories.

\if 0
\begin{figure*}[!ht]
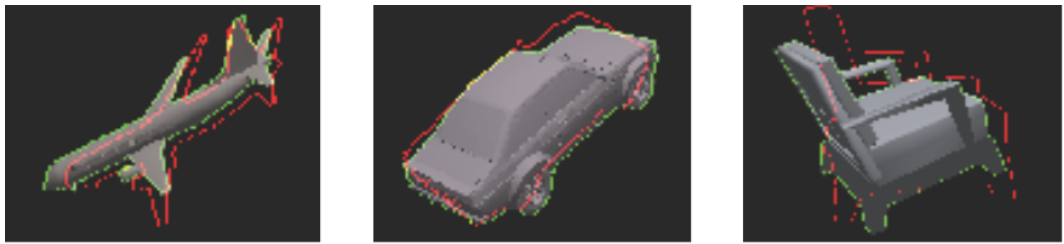

% \subfloat[airplane (init) \label{subig.u_air}]{
% \includegraphics[width=0.31\textwidth]{9_airplane_init.pdf}}
% \hfill
% \subfloat[car (init)\label{subig.u_car}]{
% \includegraphics[width=0.31\textwidth]{9_car_init.pdf}}
% \hfill
% \subfloat[chair (init)\label{subig.u_chair}]{
% \includegraphics[width=0.31\textwidth]{9_chair_init.pdf}}
% \hfill
\subfloat[airplane\label{subig.unseen_airplane}]{
\includegraphics[width=0.4\textwidth]{9_unseen_airplane_v2.pdf}}
\hfill
\subfloat[car\label{subig.unseen_car}]{
\includegraphics[width=0.4\textwidth]{9_unseen_car_v2.pdf}}
\hfill
\subfloat[chair\label{subig.unseen_chair}]{
\includegraphics[width=0.4\textwidth]{9_unseen_chair_v2.pdf}}
\hfill
\caption{Results on pose refinement of 3D models from the ModelNet dataset. These instances were not seen in training. The red and green lines represent the edges of the initial estimates and our refined poses.}
\label{fig.unseen_demo}
\end{figure*}
\fi

% \begin{figure*}[!ht]
% % \subfloat[driller \label{subig.demo_occ_ape}]{
% % \includegraphics[width=0.31\textwidth]{7_LM6d_iter4_results_1.pdf}}
% \hfill
%  \subfloat[can\label{subig.demo_occ_can}]{
%  \includegraphics[width=0.31\textwidth]{8_Occ_iter4_results_2.pdf}}
%  \hfill
%  \subfloat[ape\label{subig.demo_occ_cat}]{
%  \includegraphics[width=0.31\textwidth]{7_LM6d_iter4_results_5.pdf}}
%  \hfill
%  \caption{Demos of refined poses on the Occlusion dataset. Using the results from PoseCNN\citep{xiang2017posecnn} as initial pose.}
%  \label{fig.lm_demo}
%  \end{figure*}

%
\section{Conclusion}

In this work we introduce \dimnet, a novel framework for iterative pose matching using color images only. Given an initial 6D pose estimation of an object, we have designed a new deep neural network to directly output a \emph{relative} pose transformation that improves the pose estimate. The network automatically learns to match object poses during training. We introduce an disentangled pose representation that is also independent of the object size and the coordinate frame of the 3D object model. In this way, the network can even match poses of unseen objects, as shown in our experiments. Our method significantly outperforms state-of-the-art 6D pose estimation methods using color images only and provides performance close to methods that use depth images for pose refinement, such as using the iterative closest point algorithm. Example visualizations of our results on LINEMOD, ModelNet, T-LESS can be found here: \url{https://rse-lab.cs.washington.edu/projects/deepim}.

This work opens up various directions for future research. For instance, we expect that a stereo version of \dimnet\ could further improve pose accuracy. Furthermore, \dimnet\ indicates that it is possible to produce accurate 6D pose estimates using color images only, enabling the use of cameras that capture high resolution images at high frame rates with a large field of view, providing  estimates useful for applications such as robot manipulation.

\begin{acknowledgements}
We thank Lirui Wang at University of Washington for his contribution in this probject. This work was funded in part by a Siemens grant. We would also like to thank NVIDIA for generously providing the DGX station used for this research via the NVIDIA Robotics Lab and the UW NVIDIA AI Lab (NVAIL). This work was also Supported by National Key R\&D Program of China 2017YFB1002202, NSFC Projects 61620106005, 61325003, Beijing Municipal Sci. \& Tech. Commission Z181100008918014 and THU Initiative Scientific Research Program.
\end{acknowledgements}

\bibliographystyle{spbasic}      % basic style, author-year citations
\bibliography{egbib}

% \setlength{\tabcolsep}{4.0pt}
% \renewcommand{\arraystretch}{1.2}
% \begin{table}[t]
% \centering
% \caption{Results on the Occlusion dataset. The network is trained and tested with 4 iterations.}
% \small
% \begin{tabular}{l|c|c}
% \hline
% method 		& PoseCNN & PoseCNN+OURS\\
% \hline
% 5cm 5\degree& 1.67 & 30.93 \\
% 6D Pose  	& 28.98 & 55.04 \\
% Proj. 2D	& 15.07 & 55.05 \\
% \hline
% \end{tabular}
% \label{table.result_on_occlusion}
% \end{table}

% \section{Object Tracking on YCB dataset}
% The initial pose could be generated by means that other than a provided method. For example, once the test data is a video and images (poses) between two frames is highly correlated, we can use the estimated results from last frame as the initial pose for the current frame. We evaluated this proposal on the YCB dataset \citep{xiang2017posecnn}. During testing, PoseCNN only generate initial pose at the first frame of each video and DeepIM uses the poses from last frame as initial pose. Due to limited time, we only report our results on 3 typical categories: master chef can, cracker box and mustard bottle.
% \setlength{\tabcolsep}{4.0pt}
% \renewcommand{\arraystretch}{1.2}
% \begin{table}[t]
% \centering
% \caption{Results on Unseen objects. The network tested on }
% \small
% \begin{tabular}{l|c|c|c}
% \hline
% method 		& master chef can & cracker box & mustard\_bottle\\
% \hline
% 5cm 5\degree& 36.88 & 60.37 & \\
% 6D Pose  	& 69.68 & 73.16 & \\
% Proj. 2D	& 37.67 & 50.00 & \\
% \hline
% \end{tabular}
% \label{table.result_on_occlusion}
% \end{table}

\end{document}